%% file: main.tex
\documentclass[journal]{IEEEtran}
\usepackage{amsmath,amsfonts}
\usepackage{algorithmic}
\usepackage{algorithm}
\usepackage{array}
\usepackage[caption=false,font=normalsize,labelfont=sf,textfont=sf]{subfig}
\usepackage{textcomp}
\usepackage{stfloats}
\usepackage{url}
\usepackage{verbatim}
\usepackage{graphicx}
\usepackage{cite}
\usepackage[table]{xcolor}   
\usepackage{amssymb}    
\definecolor{car}{rgb}{0.39215686, 0.58823529, 0.96078431}
\definecolor{bicycle}{rgb}{0.39215686, 0.90196078, 0.96078431}
\definecolor{motorcycle}{rgb}{0.11764706, 0.23529412, 0.58823529}
\definecolor{truck}{rgb}{0.31372549, 0.11764706, 0.70588235}
\definecolor{other-vehicle}{rgb}{0.39215686, 0.31372549, 0.98039216}
\definecolor{person}{rgb}{1.        , 0.11764706, 0.11764706}
\definecolor{bicyclist}{rgb}{1.        , 0.15686275, 0.78431373}
\definecolor{motorcyclist}{rgb}{0.58823529, 0.11764706, 0.35294118}
\definecolor{road}{rgb}{1.        , 0.        , 1.        }
\definecolor{parking}{rgb}{1.        , 0.58823529, 1.        }
\definecolor{sidewalk}{rgb}{0.29411765, 0.        , 0.29411765}
\definecolor{other-ground}{rgb}{0.68627451, 0.        , 0.29411765}
\definecolor{building}{rgb}{1.        , 0.78431373, 0.        }
\definecolor{fence}{rgb}{1.        , 0.47058824, 0.19607843}
\definecolor{vegetation}{rgb}{0.        , 0.68627451, 0.        }
\definecolor{trunk}{rgb}{0.52941176, 0.23529412, 0.        }
\definecolor{terrain}{rgb}{0.58823529, 0.94117647, 0.31372549}
\definecolor{pole}{rgb}{1.        , 0.94117647, 0.58823529}
\definecolor{traffic-sign}{rgb}{1.        , 0.        , 0.    }

\makeatletter
\newcommand{\car@semkitfreq}{3.92}
\newcommand{\bicycle@semkitfreq}{0.03}
\newcommand{\motorcycle@semkitfreq}{0.03}
\newcommand{\truck@semkitfreq}{0.16}
\newcommand{\othervehicle@semkitfreq}{0.20}
\newcommand{\person@semkitfreq}{0.07}
\newcommand{\bicyclist@semkitfreq}{0.07}
\newcommand{\motorcyclist@semkitfreq}{0.05}
\newcommand{\road@semkitfreq}{15.30}  %
\newcommand{\parking@semkitfreq}{1.12}
\newcommand{\sidewalk@semkitfreq}{11.13}  %
\newcommand{\otherground@semkitfreq}{0.56}
\newcommand{\building@semkitfreq}{14.1}  %
\newcommand{\fence@semkitfreq}{3.90}
\newcommand{\vegetation@semkitfreq}{39.3}  %
\newcommand{\trunk@semkitfreq}{0.51}
\newcommand{\terrain@semkitfreq}{9.17} %
\newcommand{\pole@semkitfreq}{0.29}
\newcommand{\trafficsign@semkitfreq}{0.08}
\newcommand{\semkitfreq}[1]{{\csname #1@semkitfreq\endcsname}}

\newcommand\crule[3][black]{\textcolor{#1}{\rule{#2}{#3}}}
\definecolor{nvcolor}{RGB}{119,185,0}
\definecolor{roadcolor}{RGB}{234,51,246}
\definecolor{sidewalkcolor}{RGB}{68,8,72}
\definecolor{parkingcolor}{RGB}{241,156,249}
\definecolor{othergroundcolor}{RGB}{160,32,76}
\definecolor{buildingcolor}{RGB}{246,202,69}
\definecolor{carcolor}{RGB}{111,149,238}
\definecolor{truckcolor}{RGB}{74,32,172}
\definecolor{bicyclecolor}{RGB}{136,227,242}
\definecolor{motorcyclecolor}{RGB}{37,59,146}
\definecolor{othervehiclecolor}{RGB}{96,81,242}
\definecolor{vegetationcolor}{RGB}{79, 173, 50}
\definecolor{trunkcolor}{RGB}{126, 65, 22}
\definecolor{terraincolor}{RGB}{171, 238, 105}
\definecolor{personcolor}{RGB}{234, 60, 49}
\definecolor{bicyclistcolor}{RGB}{234, 66, 195}
\definecolor{motorcyclistcolor}{RGB}{138, 42, 90}
\definecolor{fencecolor}{RGB}{238, 128, 69}
\definecolor{polecolor}{RGB}{252, 241, 161}
\definecolor{trafficsigncolor}{RGB}{233, 51, 35}
\definecolor{other-struct.color}{RGB}{255, 150, 0}
\definecolor{other-objectcolor}{RGB}{50, 255, 255}
\definecolor{lane-markingcolor}{RGB}{150, 255, 170}
\definecolor{color1}{RGB}{176, 36, 24}
\definecolor{color2}{RGB}{0, 176, 80}
\definecolor{color3}{RGB}{0, 0, 200}
\definecolor{colorofteaser}{RGB}{176, 36, 24}

\definecolor{mask_red}{rgb}{1,0,0.8}
\newcommand{\green}[1]{\textcolor[RGB]{96,177,87}{#1}}
\newcommand{\red}[1]{\textcolor{mask_red}{#1}}

\definecolor{rblue}{rgb}{0,0.5,1}

\usepackage{multirow}

\usepackage{paralist}

\definecolor{hollywoodcerise}{rgb}{0.96, 0.0, 0.63}
\definecolor{lasallegreen}{rgb}{0.03, 0.47, 0.19}
\definecolor{hanpurple}{rgb}{0.32, 0.09, 0.98}
\definecolor{green(pigment)}{rgb}{0.0, 0.65, 0.31}
\usepackage[pagebackref=false,breaklinks=true,colorlinks,bookmarks=false]{hyperref}
\hypersetup{colorlinks=true,linkcolor={red},citecolor={hanpurple},urlcolor={magenta}} 

\usepackage{xspace}
\makeatletter
\DeclareRobustCommand\onedot{\futurelet\@let@token\@onedot}
\def\@onedot{\ifx\@let@token.\else.\null\fi\xspace}
\def\eg{\emph{e.g}\onedot} 
\def\ie{\emph{i.e}\onedot}

\makeatother

\usepackage{booktabs}

\begin{document}

\title{Offboard Occupancy Refinement with Hybrid Propagation for Autonomous Driving}

\author{Hao~Shi$^{1,*}$,
        Song~Wang$^{3,*}$,
        Jiaming~Zhang$^{4}$,
        Xiaoting~Yin$^{1}$,
        Guangming~Wang$^{5}$,
        Jianke~Zhu$^{3}$,\\Kailun~Yang$^{2}$,
        and~Kaiwei~Wang$^{1}$ 
\thanks{This work was supported in part by the National Natural Science Foundation of China (NSFC) under Grant No.~12174341 and No.~62473139, and in part by Shanghai SUPREMIND Technology Company Ltd. \textit{(Corresponding authors: Kaiwei Wang; Kailun Yang.)}}
\thanks{$^{1}$H. Shi, X. Yin, and K. Wang are with the State Key Laboratory of Extreme Photonics and Instrumentation and the National Engineering Research Center of Optical Instrumentation, Zhejiang University, Hangzhou 310027, China.}%
\thanks{$^{2}$K. Yang is with the School of Robotics and the National Engineering Research Center of Robot Visual Perception and Control Technology, Hunan University, Changsha 410082, China.}%
\thanks{$^{3}$S. Wang and J. Zhu are with the College of Computer Science and Technology, Zhejiang University, Hangzhou 310027, China.}%
\thanks{$^{4}$J. Zhang is with the Institute for Anthropomatics and Robotics, Karlsruhe Institute of Technology, 76131 Karlsruhe, Germany.}%
\thanks{$^{5}$G. Wang is with the Department of Engineering, University of Cambridge, Cambridge CB2 1PZ, U.K.}%
\thanks{$^{*}$These authors contributed equally.}%
}

\markboth{IEEE Transactions on Intelligent Transportation Systems, May~2025}%
{Shi \MakeLowercase{\textit{et al.}}: OccFiner}

\maketitle

\begin{abstract}
\input{Tex_content/abstract_R1}
\end{abstract}


\section{Introduction}
\label{sec:intro}

\begin{figure*}
    \centering
    \includegraphics[width=\textwidth]{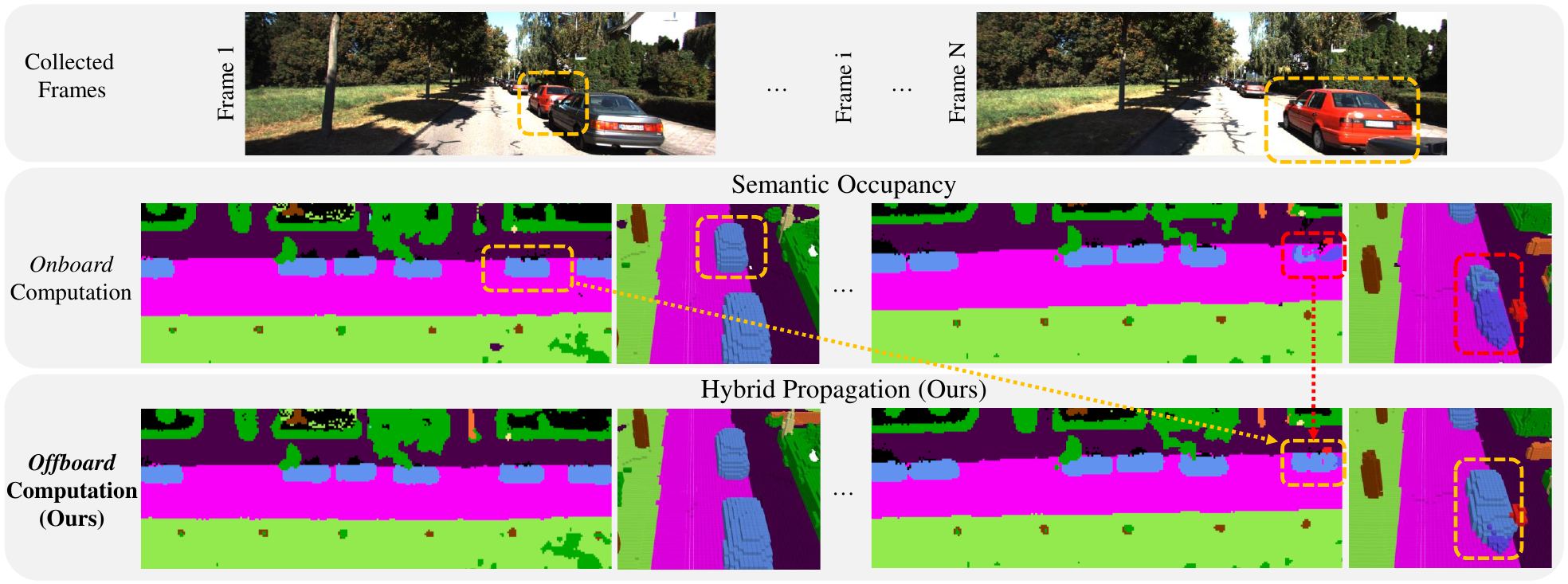}
    \vskip -2ex
    \caption{Current \emph{onboard} methods generate unreliable occupancy predictions that are inconsistent across different viewpoints. In contrast, our \emph{offboard} framework constructs a unified and multi-view consistent occupancy map with higher accuracy. 
    } 
    \label{fig_teaser}
    \vskip -2ex
\end{figure*}

\input{Tex_content/intro_R1}

\section{Related Work}
\label{sec:related_work}
\input{Tex_content/related_work_R1}

\section{Approach}
\label{sec:approach}
\input{Tex_content/approach_R1}

\section{Experiments}
\label{sec:experiments}
\input{Tex_content/exp_R1}

\vskip -1ex
\section{Conclusions and Discussion}
\label{sec:conclusions}

\input{Tex_content/conclusion_R1}

%
\bibliographystyle{IEEEtran}
\bibliography{bib_R1}

\end{document}

%% file: Tex_content/abstract_R1.tex
Vision-based occupancy prediction, also known as 3D Semantic Scene Completion (SSC), presents a significant challenge in computer vision. Previous methods, confined to onboard processing, struggle with simultaneous geometric and semantic estimation, continuity across varying viewpoints, and single-view occlusion. Our paper introduces OccFiner, a novel \emph{offboard} framework designed to enhance the accuracy of vision-based occupancy predictions. OccFiner operates in two hybrid phases: 1) a \emph{multi-to-multi local propagation network} that implicitly aligns and processes multiple local frames for correcting onboard model errors and consistently enhancing occupancy accuracy across all distances. 2) the \emph{region-centric global propagation}, focuses on refining labels using explicit multi-view geometry and integrating sensor bias, particularly for increasing the accuracy of distant occupied voxels. Extensive experiments demonstrate that OccFiner improves both geometric and semantic accuracy across various types of coarse occupancy, setting a new state-of-the-art performance on the SemanticKITTI dataset. Notably, OccFiner significantly boosts the performance of vision-based SSC models, achieving accuracy levels competitive with established LiDAR-based onboard SSC methods. Furthermore, OccFiner is the first to achieve automatic annotation of SSC in a purely vision-based approach. Quantitative experiments prove that OccFiner successfully facilitates occupancy data loop-closure in autonomous driving. Additionally, we quantitatively and qualitatively validate the superiority of the offboard approach on city-level SSC static maps. The source code will be made publicly available at \url{https://github.com/MasterHow/OccFiner}.

%% file: Tex_content/intro_R1.tex
%

Vision-based Occupancy Prediction, often referred to as Semantic Scene Completion (SSC), aims to accurately reconstruct the geometry and semantics of a 3D scene from image captures~\cite{cao2022monoscene}, providing critical location and semantic information for safe navigation in autonomous driving. Compared to LiDAR solutions, camera-based systems offer the benefits of being lightweight, cost-effective, and easy to deploy and maintain. 

However, vision-based SSC has historically lagged in accuracy compared to LiDAR due to two main limitations: 1) The depth measurement of monocular cameras lacks the precision of LiDAR systems, requiring onboard vision-based SSC methods to struggle with both geometry and semantics. 2) Onboard models are limited to processing a restricted number of local frames, impacting accuracy and multi-view continuity. 
This issue manifests as unreliable estimations and discontinuities across viewpoints in geometry and semantics. As illustrated in Fig.~\ref{fig_teaser}, most SSC efforts are focused on \textit{onboard} settings~\cite{roldao2020lmscnet,cao2022monoscene,zhang2023occformer,li2023voxformer,xiao2024instance,shepel2021occupancy,robbiano2020bayesian,li2022integrated}, \ie, data calculation and feedback occur on the vehicle itself. However, integrating data into \textit{offboard} systems is critical for identifying and rectifying potential model deficiencies, achieving data loop-closure, and enabling system self-evolution, which are essential for a wide range of intelligent transportation applications~\cite{li2023survey}. This approach solves the distribution shift between training and deployment and allows training to be scaled both safely and cost-effectively~\cite{zhang2022rethinking}. While offboard systems offer refined evaluation of onboard model performance, essential for the data loop-closure in autonomous driving~\cite{zhang2022rethinking,li2023survey,li2024data}, and well-studied in 3D detections~\cite{qi2021offboard, yang2021auto4d, fan2023once, ma2023detzero}, their potential in SSC remains under-explored.

Given such limitations of inferior performance and onboard SSC settings, we propose the first \emph{offboard} SSC framework, \textbf{OccFiner}, a novel approach that involves additional computation and refinement of on-vehicle predictions at a data center but greatly boosts the prediction reliability. The objective is to significantly enhance the reliability of vision-based SSC. By doing so, the constructed 3D SSC map can be reliably used repeatedly after a one-time low-cost acquisition with only cameras. This approach enhances auto-labeling by leveraging aggregated and diverse data from multiple sources, significantly reducing manual annotation efforts and improving labeling accuracy for large-scale 3D scene understanding~\cite{zhang2022rethinking,li2023survey,li2024data} in autonomous driving.

Specifically, we consider two primary error sources in addressing the inaccuracies of vision-based SSC: the prediction bias error introduced by the onboard model, and the measurement bias from the camera capture process. OccFiner decouples these two heterogeneous error sources and uses a two-stage hybrid process to process them separately:
1) \textit{Multi-to-Multi Local Propagation}. In the initial stage, OccFiner addresses prediction bias errors from the onboard model. We implement an error compensation model that processes inputs from off-the-shelf SSC models~\cite{cao2022monoscene,li2023voxformer,roldao2020lmscnet}. This model uniquely encodes relative spatial coordinates within a local window, facilitating the propagation of geometric and semantic cues by a newly proposed DualFlow4D transformer. This approach effectively enhances the accuracy of occupancy predictions at varying distances.
2) \textit{Region-centric Global Propagation}. The second stage focuses on measurement errors from camera captures. Here, OccFiner employs explicit multi-view geometry to register and aggregate data across the entire scene, incorporating sensor bias weighting in the process.
Compared with the simple accumulation of multi-frame averages, our region-centric voting strategy yields a more accurate SSC map. By integrating local implicit error compensation and non-local explicit information propagation from various viewpoints, OccFiner effectively addresses onboard bias, multi-view consistency, and single-view occlusions, thereby naturally enhancing the reliability and robustness of scene understanding.

\input{Tabels/abb_R1}

To the best of our knowledge, we are the first to introduce offboard refinement for SSC. Our method, termed OccFiner, enables seamless integration across various SSC models and datasets as a plug-and-play solution. For 3D semantic scene completion on the SemanticKITTI dataset~\cite{behley2019semantickitti}, OccFiner has demonstrated remarkable performance, surpassing VoxFormer by $3.67\%$ in mIoU, which makes a $26.99\%$ relative improvement and ranks first on the test leaderboard among all vision-based methods. Surprisingly, OccFiner also works well for LiDAR-based SSC~\cite{xia2023scpnet} with minor changes, leading to a new state-of-the-art of $37.82\%$ in mIoU, achieving a $9.5\%$ improvement compared with the best benchmark result.

To enhance readability, Tab.~\ref{tab:glossary} presents all abbreviations and their full forms used in this paper. 

To summarize, we deliver the following contributions:
\begin{compactitem}
\item 
We introduce OccFiner, the first dedicated offboard framework designed to enhance vision-based Semantic Scene Completion (SSC) by leveraging multi-view consistency and addressing onboard model limitations.
\item 
We propose a novel hybrid propagation strategy within OccFiner, uniquely combining multi-frame local implicit refinement (via DualFlow4D transformer) to correct prediction bias and global explicit aggregation (via region-centric voting with sensor bias) to handle measurement bias and ensure long-range consistency.
\item 
OccFiner achieves new state-of-the-art performance on the public SemanticKITTI benchmark for both camera-based and LiDAR-based SSC. Notably, our refined camera-based SSC is comparable with established LiDAR-based onboard SSC methods (\eg SSCNet-full~\cite{song2017semantic}), demonstrating substantial relative mIoU improvements (\eg, $+24.59\%$ refinement for VoxFormer on the validation set).
\item
We successfully demonstrate the first application of purely vision-based automatic annotation for SSC data loop-closure, validating OccFiner's potential for scalable data generation and system self-evolution.
\item
We validate OccFiner's effectiveness on city-level mapping, proving the benefits of the offboard approach for generating large-scale, consistent 3D semantic maps.
\end{compactitem}

%% file: Tabels/abb_R1.tex
\begin{table}[!t]
\centering
\caption{Glossary of Terms and Abbreviations.}
\label{tab:glossary}
\begin{tabular}{ll}
\toprule
\textbf{Abbreviation} & \textbf{Full Form} \\
\midrule
SSC    & Semantic Scene Completion \\
BEV    & Bird's-Eye View \\
V2X    & Vehicle-to-Everything \\
SS    & Soft Split Operation \\
LN    & Layer Normalization \\
Q    & Queries \\
K    & Keys \\
V    & Values \\
MHFA    & Multi-Head Focal Attention \\
Attn & Scaled Dot-product Attention \\
DualFlow4D & Dual-branch 4D Feature Flow Module \\
IoU    & Intersection over Union \\
mIoU    & Mean Intersection over Union \\
\textit{fov}    & Field of View \\
\bottomrule
\end{tabular}
\end{table}

%% file: Tex_content/related_work_R1.tex
\subsection{3D Semantic Scene Completion}
\label{related_work:3d_semantic_occ_prediction}

Semantic Scene Completion (SSC), also known as Semantic Occupancy Prediction, aims at concurrently estimating the geometry and semantics of a surrounding scene. 
The first effort SSCNet~\cite{song2017semantic} lays the foundation for this field by defining the SSC task and introducing a method for estimating the structure and semantics of indoor scenes from a singular depth image input. Recognizing the critical role of SSC-based perception in the realm of autonomous driving, researchers have increasingly turned their attention to its applications in outdoor scenes after the release of the large-scale outdoor benchmark SemanticKITTI~\cite{behley2019semantickitti}. 
The predominant works in this area can be broadly categorized based on the input modality: 
LiDAR-based SSC~\cite{roldao2020lmscnet,rist2021semantic_implicit,yan2021sparse_sweep,cheng2021_s3cnet,yang2021semantic_assisted,zuo2023pointocc,xia2023scpnet,yan2023spot} and 
vision-based SSC~\cite{silva2024_s2tpvformer,ma2023cotr,mei2023camera,wang2023panoocc,lyu2023_3dopformer,wei2023surroundocc,ma2023cam4docc,lu2023octreeocc,jiang2023symphonize,yu2023flashocc,li2023fb_occ, cao2022monoscene,zhang2023occformer,hou2024fastocc}.
Additionally, recent research expands into several novel areas within SSC including the world model~\cite{zheng2023occworld}, 
3D object detection~\cite{jia2023occupancydetr,tong2023scene_as_occupancy,peng2023learning_occupancy_3d,xu2023regulating,hong2024univision,zhou2023sogdet,min2024multi_camera},
multimodal parsing~\cite{li2023stereoscene,yao2023depthssc,zhang2023radocc,pan2023renderocc,miao2023occdepth,ming2024_inversematrixvt3d,xiao2024instance,liu2023fully_sparse,ming2024occfusion},
self-supervised prediction~\cite{hayler2023s4c,zhang2023occnerf,min2023occupancy_mae,huang2023selfocc,boeder2024occflownet},
open-vocabulary recognition~\cite{vobecky2023pop_3d,tan2023ovo}, 
dense top-view understanding~\cite{peng2022mass,gu2023dense_semantic_completion,lu2022cylindrical_dense_top_view}, V2X collaboration~\cite{li2023multi_robot,zhang2024v2vssc,song2024collaborative_occupancy},
as well as various benchmarks~\cite{wang2023openoccupancy,tian2023occ3d,li2023sscbench}.
However, the majority of previous studies have concentrated on methodological design within onboard settings, where observations across different viewpoints often exhibit notable discontinuities. In contrast, OccFiner shifts focus to the offboard setting, aiming to rectify errors inherent in onboard models and to aggregate long-term geometric and semantic cues. Our approach contributes to data closure in autonomous driving, addressing a crucial aspect that has been largely overlooked in previous SSC research.

\subsection{Offboard 3D Perception}
\label{related_work:offboard_3d_perception}
Perception in autonomous driving scenarios necessitates extensive data annotation and training~\cite{geiger2012we,caesar2020nuscenes, sun2020scalability}. Beyond the real-time `onboard' processing performed directly on the vehicle, `offboard' perception involves processing collected sensor data after the driving mission, typically within a data center or cloud environment. This paradigm leverages significantly larger computational resources and allows for the aggregation of data from multiple vehicles or extensive driving logs. Key goals of offboard systems include identifying and rectifying potential onboard model deficiencies, achieving data loop-closure~\cite{li2023survey, zhang2022rethinking, li2024data}, enabling large-scale automatic annotation, and facilitating system self-evolution~\cite{li2023survey}, which are crucial for scaling autonomous driving capabilities safely and cost-effectively~\cite{zhang2022rethinking}.
Since LiDAR can provide accurate range information in 3D space, researchers explore its sequence learning in semantic segmentation~\cite{wang2022meta_rangeseg, aygun20214d}, object detection~\cite{najibi2022motion, park2022time}, and tracking~\cite{pang2021model} to obtain better performance. 
The sequential characteristics of LiDAR scans are further studied in offboard 3D object detection~\cite{qi2021offboard, yang2021auto4d, fan2023once, ma2023detzero} to reduce annotation costs. 
3DAL~\cite{qi2021offboard} formulates the problem of offboard 3D object detection and proposes an \textit{object-centric} auto-labeling detector.
CTRL~\cite{fan2023once} adheres to the principle of \textit{track-centric} and elevates the accuracy of auto-labeling, surpassing manual annotation in some scenarios. 
While these aforementioned methods demonstrated the value of the offboard approach, they were primarily designed for LiDAR point clouds and focused on object detection or tracking. The challenge of applying offboard refinement to dense, vision-based Semantic Scene Completion (SSC), which involves reconstructing both geometry and semantics voxel-wise, remained under-explored. Vision-based approaches like MV-Map~\cite{xie2023mv} provided \textit{region-centric} fusion for high-definition maps but only processed limited traffic elements on the bird's-eye view plane.
To address these gaps and harness the benefits of offboard processing for dense scene understanding from cameras, we design the first hybrid propagation pipeline (OccFiner) specifically for offboard SSC.
OccFiner can process the SSC predictions from either cameras or LiDAR and provide high-quality occupancy information for autonomous vehicles, which holds significant importance in auto-labeling and data closure in pure visual perception.

%% file: Tex_content/approach_R1.tex
\subsection{Overview}
\label{approach:overview}

\begin{figure*}
    \centering
    \includegraphics[width=0.95\textwidth]{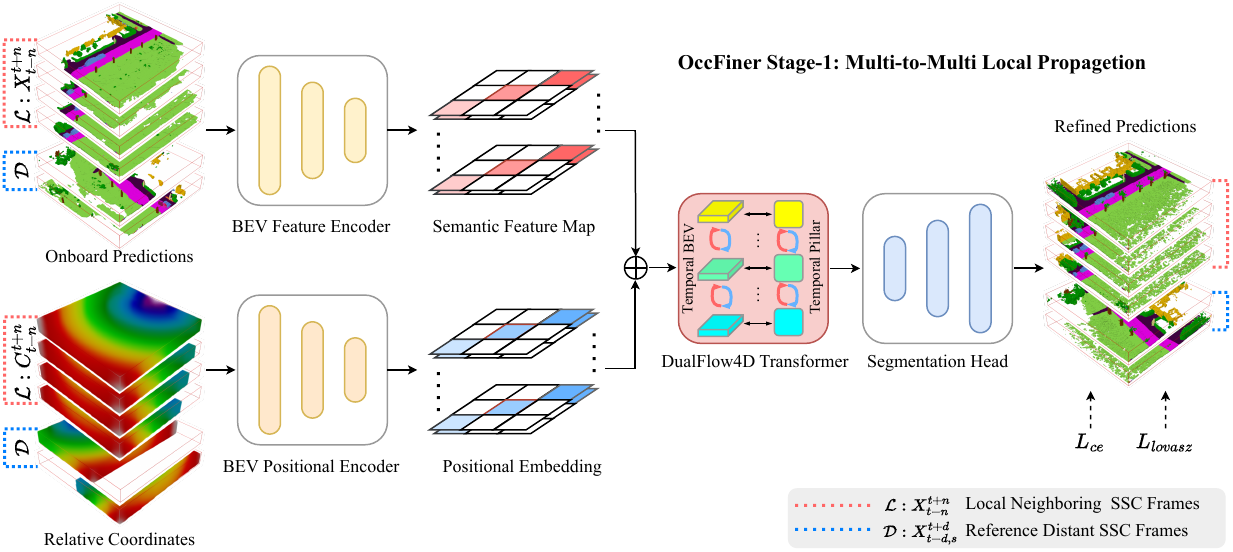}
    \caption{\textbf{Overview of the proposed multi-to-multi local propagation network (OccFiner Stage 1).} 
    It accepts multiple onboard predictions and relative coordinates as input (multi-input) and simultaneously generates refined predictions for multiple frames (multi-output). This network adeptly executes error compensation and facilitates implicit local propagation, improving SSC quality and temporal consistency across various distances.} 
    \label{fig_stage1}
\end{figure*}

The OccFiner framework, designed for offboard Semantic Scene Completion (SSC), operates in two distinct stages. The first stage focuses on compensating for errors in onboard model predictions, employing a multi-to-multi local propagation network to process SSC model outputs across a series of frames, utilizing both close-range and distant frames as references. This stage effectively merges relative spatial coordinates with semantic features, employing transformer-based methods for spatio-temporal integration. The second stage of OccFiner emphasizes global semantic and geometric aggregation. It transforms refined voxel labels into semantic point clouds and applies relative poses over extended periods for precise coordinate adjustments, incorporating sensor measurement characteristics for voxel voting. This dual-stage strategy effectively combines local implicit feature propagation with global explicit semantic aggregation, forming a robust hybrid system for offboard SSC refinement.

\subsection{Multi-to-Multi Local Propagation Network}
\label{approach:local_prop}

\noindent \textbf{Problem Formulation:}
Let $X^T := \{X_1, X_2, ..., X_T \}$ be an onboard SSC inference sequence of height $H$, width $W$, depth $Z$, and $T$ frames in length.
$C^T := \{C_1, C_2, ..., C_T\}$ denotes the corresponding frame-wise relative coordinates, which can be calculated by:    
\begin{equation}
   \mathbf{c}_{i} = (T_{li}^{cam})^{-1} (T_{cam_{t}}^{world})^{-1} T_{cam_{i}}^{world} T_{li}^{cam} \mathbf{x}_{i}, 
   \label{equ:relative_coords}
\end{equation} 
\noindent
where each vertex $\mathbf{x}_{i} \in X_i$ is projectively associated with a relative vertex $\mathbf{c}_{i} \in C_i$.
For each relative coordinate $C_i$, it represents the mapping relationship between the three-dimensional coordinates of the current frame $X_i$ and the coordinates of the pivot frame $X_t$. 
We formulate the occupancy refinement as a task that takes the onboard ($X^T, C^T$) pairs as input and reconstruct the SSC labels $Y^T = \{Y_1, Y_2, ..., Y_T\}$.
Specifically, we propose to learn a mapping function from onboard inferences $X^T$ to the output $\hat{Y}^T := \{\hat{Y}_1, \hat{Y}_2, ..., \hat{Y}_T\}$, 
such that the conditional distribution of the real data $p(Y^T|X^T)$ can be approximated by the distribution of generated data $p(\hat{Y}^T | X^T)$. 

We conceptualize the task as a `multi-to-multi' prediction problem: the network takes multiple input frames ($X^{T}$) and simultaneously refines multiple output frames ($\hat{Y}^{T}$) within that sequence in a single feed-forward pass. This approach is chosen specifically to leverage temporal context and enforce consistency across the refined sequence, unlike methods that process frames individually or refine only a single central frame.
The intuition is that regions obscured in the current viewpoint are likely to become visible in frames captured from a distance, particularly in scenarios involving large occlusions or when the vehicle is moving at a slower pace. In the context of multi-frame offboard SSC, it is more effective to address gaps in a target frame by leveraging content from the entire scene sequence, incorporating information from both neighboring and distant frames as conditional inputs. This approach relies on the Markov assumption~\cite{hausman1999independence} for simplification, which allows us to express the refinement process as:
\begin{equation}
p(\hat{Y}^T|X^T) = \prod_{t=1}^{T} p(\hat{Y}_{t-n}^{t+n}| X_{t-n}^{t+n}, X_{t-d, s}^{t+d}),
\label{equ:refine_process}
\end{equation}
where $X_{t-n}^{t+n}$ represents a short SSC clip of neighboring frames centered around time $t$ with a temporal radius $n$. $X_{t-d, s}^{t+d}$ denotes distant frames that are randomly sampled from a distant scene range $d$ at a rate of $s$. This selection of distant frames typically encompasses most of the key moments in possible viewpoints, effectively conveying long-term geometric and semantic cues of the scene. 

Within this framework, offboard SSC models are tasked with not only maintaining temporal consistency in neighboring frames but also ensuring that the refined frames are coherent with the broader narrative of the scene sequence. This approach aims to enhance scene understanding by integrating both local and global temporal perspectives in the refinement process.

\noindent \textbf{Network Design:}
The overview of the proposed multi-to-multi local propagation network is shown in Fig.~\ref{fig_stage1}. As indicated in Eq.~\ref{equ:refine_process}, OccFiner takes both neighboring local SSC frames $X_{t-n}^{t+n}$, distant reference SSC frames $X_{t-d, s}^{t+d}$, and relative coordinate $\{ C_{t-n}^{t+n}, C_{t-d, s}^{t+d} \}$ as conditions, to propagate across all input frames simultaneously. 
Specifically, OccFiner comprises four integral components, 
including a frame-level Bird's-Eye View (BEV) feature encoder,
a parallel BEV positional encoder, a series of spatial-temporal transformers, and a frame-level segmentation head. The frame-level BEV feature encoder is architecturally configured with multiple 2D convolutions along the horizontal and vertical dimensions (X, Y), thus converting the height dimension (Z) into a feature dimension. Mirroring this, the frame-level BEV positional encoder shares the same architectural blueprint with the feature encoder; such a design choice aids in better aligning the positional and feature information, as both are processed through a similar computational pathway. 
At the heart of OccFiner lie the DualFlow4D transformers,
tasked with learning and applying joint spatial-temporal transformations. These transformers are specifically tailored to address uncertainties and occlusions within the deep encoding space.
\begin{figure}
    \centering
    \includegraphics[width=0.5\textwidth]{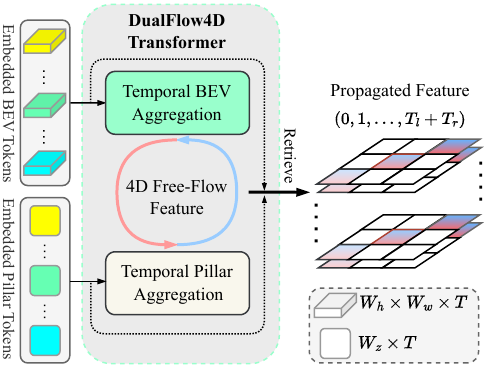}
    \caption{\textbf{Our proposed DualFlow4D transformer block.}
    It engages spatiotemporal propagation within BEV space and vanilla attention to pillar tokens. This dual approach enables effective matching and flow of semantic and geometric cues for comprehensive scene understanding.
    } 
    \label{fig:4d_trans}
\end{figure}

\noindent \textbf{DualFlow4D Transformer:}
To propagate high-fidelity features in each SSC frame, multi-layer DualFlow4D transformers are designed to search for coherent from all the encoded features. 
Specifically, we propose to first search by a multi-head soft-patch-based focal attention module along BEV and temporal dimensions. 
We use $f^T = \{f_1, f_2, ..., f_T\}$ to denote the encoded BEV features, where $f_i \in \mathbb{R} ^{h\times w\times c}$, and $T$ is the total of local and reference frames. Concurrently, $p_{l} \in \mathbb{R}^{T_{l} \times h \times w \times c}$ and $p_{r} \in \mathbb{R}^{T_{r} \times h \times w \times c}$ encapsulate the corresponding encoded local and reference relative coordinates, where $T_{l}$ and $T_{r}$ are the time dimension of local frames and reference frames.
Let ${\rm SS}(\cdot)$ denote the soft split operation, which reshapes features into overlapping patches. Let $\oplus$ denote the feature concatenation along the channel dimension.
We first embed the features and coordinates into overlapped patches $P \in \mathbb{R}^{(T_{l}+T_{r}) \times W_{h} \times W_{w} \times C_{e}}$ as follows:
\begin{equation}
\label{equ:soft_split}
    \begin{aligned}
    P = {\rm SS}(({f}_{l} \oplus f_{r}) + ({p}_{l} \oplus p_{r})).
    \end{aligned}
\end{equation}
Here, $f_l$ and $f_r$ represent the features from local and reference frames, respectively. $W_{h}$ and $W_{w}$ are the spatial dimensions of the embedded tokens, and $C_{e}$ is the embedding channel dimension.
The overlapped position aggregates a piece of information from different tokens, contributing to smoother semantic patch boundaries and enlarging its receptive field by fusing cues from neighboring patches.

Instead of the vanilla vision transformer~\cite{dosovitskiy2020image}, we use the focal transformer mechanism~\cite{yang2021focal} to efficiently aggregate information from local and non-local neighboring tokens and propagate high-fidelity BEV features. 
Let $P$ be the sequence of embedded patches from Eq.~(\ref{equ:soft_split}). We first apply Layer Normalization (${\rm LN}$) to $P$. Then, a linear projection layer, denoted as $\mathcal{P}_{qkv}$, maps the normalized tokens to Queries ($Q$), Keys ($K$), and Values ($V$):
\begin{equation}
\label{equ:focal_attention_1}
    \begin{aligned}
    & Q, K, V = \mathcal{P}_{qkv}({\rm LN}(P)). \\
    \end{aligned}
\end{equation}
\noindent These $Q$, $K$, and $V$ are then processed by the Multi-Head Focal Attention module (${\rm MHFA}$). The resulting aggregated feature for the BEV space, $Z_{bev}$, is obtained by adding the original input $P$ (residual connection):
\begin{equation}
\label{equ:focal_attention_2}
    \begin{aligned}
    & Z_{bev} = {\rm MHFA}(Q, K, V) + P. \\
    \end{aligned}
\end{equation}
\noindent For simplicity in notation, we omit the explicit time dimension here.
The propagation described by Eq.~(\ref{equ:focal_attention_2}) operates primarily within the temporal BEV space. To ensure effective feature propagation along the height dimension (Z-axis), we introduce an additional attention mechanism. This focuses on `pillars,' where each pillar represents a voxel stack along the height axis at a specific BEV location ($x, y$).
We reshape the features to obtain pillar-specific Queries ($Q_{z}$), Keys ($K_{z}$), and Values ($V_{z}$), typically with dimensions like $\mathbb{R}^{W_z \times T \times c_{e}}$. Let $d$ be the dimension of the key vectors. We compute the attention output along the pillars, $Z_{z}$, using standard scaled dot-product attention (${\rm Attn}$):
\begin{equation}
\label{equ:height_att_01}
    \begin{aligned}
    Z_{z} := {\rm Attn}(Q_{z}, K_{z}, V_{z}) = {\rm Softmax}(\frac{Q_{z} K_{z}^{\top}V_{z}}{\sqrt{d}}).
    \end{aligned}
\end{equation}
\noindent Having computed the BEV-focused features $Z_{bev}$ and the pillar-focused features $Z_{z}$, we combine them to form the final output of the DualFlow4D block. Let $\mathcal{P}_{z}$ and $\mathcal{P}_{bev}$ be linear projection layers specific to the pillar and BEV branches, respectively. The final 4D propagated feature, $Z_{4d}$, is obtained by projecting each branch's output and summing them:
\begin{equation}
\label{equ:height_att_02}
    \begin{aligned}
    Z_{4d} = \mathcal{P}_{z}(Z_z) + \mathcal{P}_{bev}(Z_{bev}).
    \end{aligned}
\end{equation}
\noindent Note that we omit the multi-head dimension in the notation for simplicity. This 4D dual-branch design promotes learning coherent spatial-temporal transformations by explicitly processing both BEV information flow ($Z_{bev}$) and height dimension information flow ($Z_z$) along the pillar-shape voxels.

\begin{figure*}
    \centering
    \includegraphics[width=0.95\textwidth]{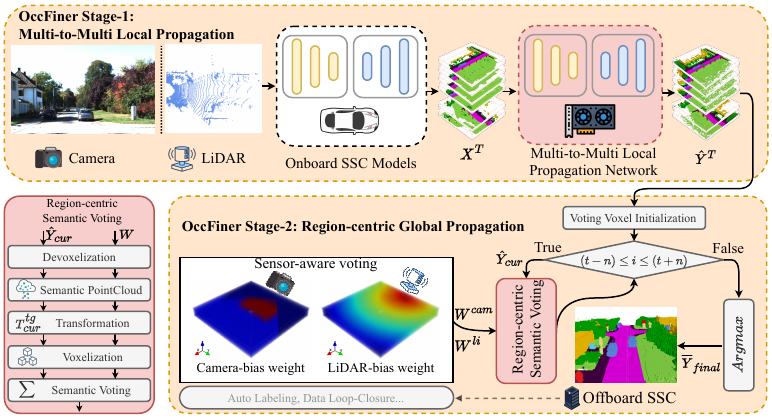}
    \caption{\textbf{Overview of the OccFiner two-stage pipeline.}
    Stage 1 performs Multi-to-Multi Local Propagation to refine onboard SSC predictions ($X^T$) into $\hat{Y}^T$. Stage 2 then performs Region-centric Global Propagation on $\hat{Y}^T$, leveraging multi-view geometry and sensor-aware voting to generate the final consistent and accurate offboard SSC map, which can be further applied to auto labeling or data loop-closure.
    } 
    \label{fig_stage2}
\end{figure*}

\subsection{Region-centric Global Propagation}
\label{approach:global_prop}

In the initial stage, the OccFiner framework engages in multi-to-multi local propagation for offboard SSC (Semantic Scene Completion) refinement. 
This stage, however, is limited by the input multi-frame window size, leading to a necessity for global propagation across the entire scene sequence. 
Notably, the local propagation process uniformly enhances SSC accuracy regardless of the various distances, a key aspect of global propagation's effectiveness. 
Additionally, OccFiner utilizes relative coordinates in the initial stage, ensuring the implicit alignment of multiple frames and maintaining the contextual integrity of each frame's semantic content. 
This approach, which prevents truncation, is pivotal for learning-based error compensation. 
The subsequent stage involves geometric explicit registration voting, covering the entire scene sequence, and forming a cohesive and hybrid strategy within the OccFiner framework.
In the global propagation stage, OccFiner focuses on addressing the measurement limitations of camera and LiDAR sensors and aggregates the independent SSC predictions into a multi-view consistent prediction for each region. 
OccFiner adeptly integrates sensor-aware weight biases with vanilla registration methods, aligning the refinement process with the unique attributes of each sensor type. 
Next, we delve into the details of vanilla registration and enhanced sensor-aware registration.

\noindent \textbf{Vanilla Registration:}
As illustrated in Fig.~\ref{fig_stage2}, it aggregates independent SSC predictions from $2n+1$ frames into a multi-view consistent prediction for each region, where $n$ is the temporal window radius. 
This involves the aggregation of per-frame semantics and geometry $\{ \hat{Y}_{i} \}_{i=1}^{2n+1}$ into a refined SSC map. For a given target voxel index $\mathbf{y}$ within a 3D region of target $\hat{Y}_{tg}$,
the process starts with devoxelizing the current SSC voxel $\hat{Y}_{cur}$.
Let $(o_x, o_y, o_z)$ be the coordinate origin of the voxel grid, $\text{dv} = (\text{dv}_x, \text{dv}_y, \text{dv}_z)$ be the size of each voxel, and $c$ be a semantic class label.
For each occupied voxel $\hat{Y}_{ijk} = c$ at index $(i,j,k)$, we derive a point in the semantic point cloud $\mathbf{P}_{cur}$ in the current LiDAR coordinate system as follows:
\begin{equation}
\label{equ:voting_01}
    \begin{aligned}
    \mathbf{P}_{cur}(x, y, z, c) &= \left\{ \left( o_x + i \cdot \text{dv}_x, o_y + j \cdot \text{dv}_y, \right. \right. \\
    &\left. \left. o_z + k \cdot \text{dv}_z, c \right) \,|\, \hat{Y}_{ijk} = c \right\}.
    \end{aligned}
\end{equation}

\noindent Next, this point cloud is then transformed relative to the target LiDAR coordinate system with poses $\{ {T}_{cur}^{tg} \}_{tg-n}^{tg+n}$:
\begin{equation}
\label{equ:voting_02}
    \begin{aligned}
    \mathbf{P}_{cur}^{tg} = {T}_{cur}^{tg} \mathbf{P}_{cur},
    \end{aligned}
\end{equation}
\noindent followed by voxelization and addition in the voting voxel. 
The final step involves voting for 
the final semantic voxel output $\overline{Y}_{final}$:
\begin{equation}
\label{equ:voting_03}
    \begin{aligned}
    \overline{Y}_{final} = \underset{c}{\mathrm{argmax}} \left( \sum_{cur=tg-n}^{tg+n} \text{Voxelize}(\mathbf{P}_{cur}^{tg}, c) \right),
    \end{aligned}
\end{equation}
\noindent where the $\mathrm{argmax}$ function selects the class 
$c$ with the highest aggregated value in the voting voxel, determining the final semantic voxel.

\noindent \textbf{Incorporating with Sensor Bias:}
While vanilla registration averages SSC across multiple frames, it overlooks the differential weighting of different locations. 
To address this, sensor-aware weighting is introduced, tailored to the distinct characteristics of cameras and LiDARs. 
This approach is informed by the measurement characteristics of measurement devices like cameras and LiDARs. 
Camera systems are constrained by their field of view, with points within the viewing frustum yielding higher accuracy compared to those outside. 
Additionally, the occlusion of distant points by nearer ones necessitates lower voting weights for the former. 
Conversely, LiDAR systems exhibit increased sparsity and depth measurement errors at longer distances. 
Therefore, for cameras, points within the field of view ($fov$) and closer ranges are given higher voting weights.
Let $F_{cam}(x_{c}, y_{c}, z_{c})$ be a boolean function indicating if a voxel is within the camera frustum, defined as:
\begin{equation}
\label{equ:weight_00}
    \begin{aligned}
    F_{cam}(x_{c}, y_{c}, z_{c}) &= \left( |\text{atan2}(x_c, z_c)| \leq \frac{fov_{w}}{2} \right) \\
    &\land \left( |\text{atan2}(y_c, z_c)| \leq \frac{fov_{h}}{2} \right) \\
    &\land (z_{c} > 0),
    \end{aligned}
\end{equation}
\noindent where $(x_c, y_c, z_c)$ denote the voxel indices in the camera coordinate system.
Let $B(\mathbf{x})$ be a boolean function indicating if a voxel vertex $\mathbf{x}$ is inside a predefined bounding box (used to prioritize closer regions). We define three weight levels: high ($w_{\text{high}}$), medium ($w_{\text{med}}$), and low ($w_{\text{low}}$). The final camera voting weight $W_{cam}(\mathbf{x})$ for a voxel vertex $\mathbf{x}$ is assigned as:
\begin{equation}
W_{cam}(\mathbf{x}) = \begin{cases}
    w_{\text{high}}, & \text{if $(F_{cam} \land B)(\mathbf{x}) > 0$}; \\
    w_{\text{med}}, & \text{if $ [ F_{cam}(\mathbf{x}) > 0 ]$ $\land$ $ [ B(\mathbf{x}) < 0 ] $}; \\
    w_{\text{low}}, & \text{otherwise}.
\end{cases}
\end{equation}
Conversely, for LiDARs, accuracy typically decreases with distance due to sparsity and measurement errors.
Let $r$ be the radial distance of a voxel from the LiDAR origin, $R$ be the maximum considered range, and $w_{\text{max}}$ and $w_{\text{min}}$ be the maximum and minimum voting weights. The voting weight $W_{li}(r)$ linearly attenuates with increasing distance $r$:
\begin{equation}
\label{equ:weight_02}
    \begin{aligned}
    W_{li}(r) = w_{\text{max}} - (w_{\text{max}} - w_{\text{min}}) \times \frac{r}{R}.
    \end{aligned}
\end{equation}

\input{Tabels/semantic-kitti-val-cam}

Consequently, the OccFiner framework adopts a hybrid dual-stage strategy: the first stage addresses onboard model errors, enhancing prediction accuracy regardless of the range, and the second stage adjusts for the physical limitations of cameras and LiDARs, further improving the offboard accuracy.

\subsection{Supervision}
\label{approach:supervision}
The multi-to-multi local propagation network is trained end-to-end using the ground-truth labels corresponding to the input frames. The overall loss function for Semantic Scene Completion ($\mathcal{L}_{ssc}$) comprises two commonly used components. The first is the standard voxel-wise Cross-Entropy loss, denoted as ($\mathcal{L}_{ce}$). The second is the Lov\'asz loss~\cite{berman2018lovasz}, denoted as $\mathcal{L}_{lovasz}$, which is designed as a surrogate for directly optimizing the intersection-over-union (IoU) metric crucial for segmentation tasks. The total SSC loss is the sum of these two components:
\begin{equation}
\mathcal{L}_{ssc}=\mathcal{L}_{ce}+\mathcal{L}_{lovasz}.
\end{equation}

%% file: Tabels/semantic-kitti-val-cam.tex
\begin{table*}[!t]
    \scriptsize
    \setlength{\tabcolsep}{0.0035\linewidth}    
    \caption{\textbf{Camera-based semantic scene completion results on the SemanticKITTI validation set}~\cite{behley2019semantickitti}\textbf{.} * represents these methods are adapted for the RGB inputs, which are implemented and reported in MonoScene~\cite{cao2022monoscene}. $\dagger$ represents the reproduced result from~\cite{huang2023tpvformer}. 
    }
    \resizebox{1.0\textwidth}{!}
{
    \newcommand{\classfreq}[1]{{~\tiny(\semkitfreq{#1}\%)}}  %
    \centering
    \begin{tabular}{l|c c c c c c c c c c c c c c c c c c c|c c}
            \toprule
    & & \multicolumn{20}{c}{} \\
    Method 
    & \rotatebox{90}{\textcolor{car}{$\blacksquare$} car\classfreq{car}}
    & \rotatebox{90}{\textcolor{bicycle}{$\blacksquare$} bicycle\classfreq{bicycle}}
    & \rotatebox{90}{\textcolor{motorcycle}{$\blacksquare$} motorcycle\classfreq{motorcycle}}
    & \rotatebox{90}{\textcolor{truck}{$\blacksquare$} truck\classfreq{truck}}
    & \rotatebox{90}{\textcolor{other-vehicle}{$\blacksquare$} other-vehicle\classfreq{othervehicle}}
    & \rotatebox{90}{\textcolor{person}{$\blacksquare$} person\classfreq{person}}
    & \rotatebox{90}{\textcolor{bicyclist}{$\blacksquare$} bicyclist\classfreq{bicyclist}}
    & \rotatebox{90}{\textcolor{motorcyclist}{$\blacksquare$} motorcyclist\classfreq{motorcyclist}}
    & \rotatebox{90}{\textcolor{road}{$\blacksquare$} road\classfreq{road}}
    & \rotatebox{90}{\textcolor{parking}{$\blacksquare$} parking\classfreq{parking}}
    & \rotatebox{90}{\textcolor{sidewalk}{$\blacksquare$} sidewalk\classfreq{sidewalk}}
    & \rotatebox{90}{\textcolor{other-ground}{$\blacksquare$} other-ground\classfreq{otherground}}
    & \rotatebox{90}{\textcolor{building}{$\blacksquare$} building\classfreq{building}}
    & \rotatebox{90}{\textcolor{fence}{$\blacksquare$} fence\classfreq{fence}}
    & \rotatebox{90}{\textcolor{vegetation}{$\blacksquare$} vegetation\classfreq{vegetation}}
    & \rotatebox{90}{\textcolor{trunk}{$\blacksquare$} trunk\classfreq{trunk}}
    & \rotatebox{90}{\textcolor{terrain}{$\blacksquare$} terrain\classfreq{terrain}}
    & \rotatebox{90}{\textcolor{pole}{$\blacksquare$} pole\classfreq{pole}}
    & \rotatebox{90}{\textcolor{traffic-sign}{$\blacksquare$} traffic-sign\classfreq{trafficsign}}
    & IoU & mIoU\\
    \midrule
        LMSCNet*~\cite{roldao2020lmscnet} & 18.33 & 0.00 & 0.00 & 0.00 & 0.00 & 0.00 & 0.00 & 0.00 & 40.68 & 4.38 & 18.22 & 0.00 & 10.31 & 1.21 & 13.66 & 0.02 & 20.54 & 0.00 & 0.00 & 28.61 & 6.70 \\ 
        
        3DSketch*~\cite{chen20203d} & 18.59 & 0.00 & 0.00 & 0.00 & 0.00 & 0.00 & 0.00 & 0.00 & 41.32 & 0.00 & 21.63 & 0.00 & 14.81 & 0.73 & 19.09 & 0.00 & 26.40 & 0.00 & 0.00 & 33.30 & 7.50 \\

        AICNet*~\cite{li2020anisotropic} & 14.71 & 0.00 & 0.00 & 4.53 & 0.00 & 0.00 & 0.00 & 0.00 & 43.55 & 11.97 & 20.55 & 0.07 & 12.94 & 2.52 & 15.37 & 2.90 & 28.71 & 0.06 & 0.00 & 29.59 & 8.31\\

        JS3C-Net*~\cite{yan2021sparse_sweep} & 24.65 & 0.00 & 0.00 & 4.41 & 6.15 & 0.67 & 0.27 & 0.00 & 50.49 & 11.94 & 23.74 & 0.07 & 15.03 & 3.94 & 18.11 & 4.33 & 26.86 & 3.77 & 1.45 & 38.98 & 10.31\\

        OccFormer~\cite{zhang2023occformer} & 25.09 & 0.81 & 1.19 & 25.53 & \textbf{8.52} & 2.78 & \textbf{2.82} & 0.00 & 58.85 & 19.61 & 26.88 & 0.31 & 14.40 & 5.61 & 19.63 & 3.93 & 32.62 & 4.26 & 2.86 & 36.50 & 13.46\\
        
        MonoScene$\dagger$~\cite{cao2022monoscene} & 23.26 & 0.61 & 0.45 & 6.98 & 1.48 & 1.86 & 1.20 & 0.00 & 56.52 & 14.27 & 26.72 & 0.46 & 14.09 & 5.84 & 17.89 & 2.81 & 29.64 & 4.14 & 2.25 & 36.86 & 11.08\\

        TPVFormer~\cite{huang2023tpvformer} & 23.81 & 0.36 & 0.05 & 8.08 & 4.35 & 0.51 & 0.89 & 0.00 & 56.50 & 20.60 & 25.87 & 0.85 & 13.88 & 5.94 & 16.92 & 2.26 & 30.38 & 3.14 & 1.52 & 35.61 & 11.36\\

        VoxFormer~\cite{li2023voxformer} & 27.01 & 1.05 & 0.47 & 9.90 & 4.64 & 1.51 & 0.85 & 0.00 & 54.67 & 18.55 & 27.35 & 0.41 & 19.65 & 8.52 & 26.18 & 6.58 & 32.39 & 8.69 & 4.71 & 44.16 & 13.33 \\ 

        \midrule
        Average Fusion (Vox) & 30.35 & 0.88 & 0.41 & 12.92 & 3.99 & 1.60 & 0.86 & 0.00 & 57.60 & 18.53 & 29.44 & 0.28 & 21.83 & 10.50 & 29.24 & 7.44 & 34.61 & \textbf{9.79} & \textbf{5.54} & 46.03 & 14.52 \\ 
        
        \midrule
        \rowcolor{gray!20}
        OccFiner (Mono) & 29.12 & 0.41 & \textbf{1.41} & 21.19 & 7.85 & 3.97 & 2.28 & 0.00 & 56.62 & 19.64 & 27.62 & 1.48 & 15.07 & 7.18 & 20.89 & 3.89 & 31.79 & 3.92 & 2.39 & 37.42 & 13.45 \\
        \rowcolor{gray!20}
        \textit{w.r.t. MonoScene} & \red{+5.86} & \green{-0.20} & \red{+0.96} & \red{+14.21} & \red{+6.37} & \red{+2.11} & \red{+1.08} & - & \red{+0.10} & \red{+5.37} & \red{+0.90} & \red{+1.02} & \red{+0.98} & \red{+1.34} & \red{+3.00} & \red{+1.08} & \red{+2.15} & \green{-0.22} & \red{+0.14} & \red{+0.56} & \red{+2.37} \\

        \midrule
        \rowcolor{gray!20}
        OccFiner (TPV) & 29.19 & 0.69 & 0.00 & 22.73 & 7.05 & 0.95 & 1.40 & 0.00 & 57.15 & 22.78 & 28.35 & \textbf{7.93} & 15.10 & 6.87 & 19.88 & 3.09 & 31.02 & 3.70 & 1.92  & 36.54 & 13.30 \\
        \rowcolor{gray!20}
        \textit{w.r.t. TPVFormer} & \red{+5.38} & \red{+0.33} & \green{-0.05} & \red{+14.65} & \red{+2.70} & \red{+0.44} & \red{+0.51} & - & \red{+0.65} & \red{+2.18} & \red{+2.48} & \red{+7.08} & \red{+1.22} & \red{+0.93} & \red{+2.96} & \red{+0.83} & \red{+0.64} & \red{+0.56} & \red{+0.40} & \red{+0.93} & \red{+1.94} \\

        \midrule    %
        \rowcolor{gray!20}

        OccFiner (Vox) & \textbf{36.78} & \textbf{1.73} & 0.29 & \textbf{32.96} & 5.67 & \textbf{4.06} & 1.12 & 0.00 & \textbf{66.34} & \textbf{27.05} & \textbf{35.06} & 0.35 & \textbf{25.46} & \textbf{11.79} & \textbf{32.82} & \textbf{9.87} & \textbf{40.56} & 8.02 & 3.58 & \textbf{47.86} & \textbf{18.09} \\ %
        \rowcolor{gray!20}
        \textit{w.r.t. VoxFormer} & \red{+9.77} & \red{+0.68} & \green{-0.18} & \red{+22.05} & \red{+1.03} & \red{+2.55} & \red{+0.27} & - & \red{+11.67} & \red{+8.50} & \red{+7.71} & \green{-0.06} & \red{+5.81} & \red{+3.27} & \red{+6.64} & \red{+3.29} & \red{+8.17} & \green{-0.67} & \green{-1.13} & \red{+3.70} & \red{+4.76} \\

        \bottomrule
    \end{tabular}
    }
    \\
    \label{table:kitti_val_perf}
\end{table*}

%% file: Tex_content/exp_R1.tex
\input{Tabels/semantic-kitti-val-lidar-short-R1}

\begin{figure*}[!t]
    \centering
    \includegraphics[width=\textwidth]{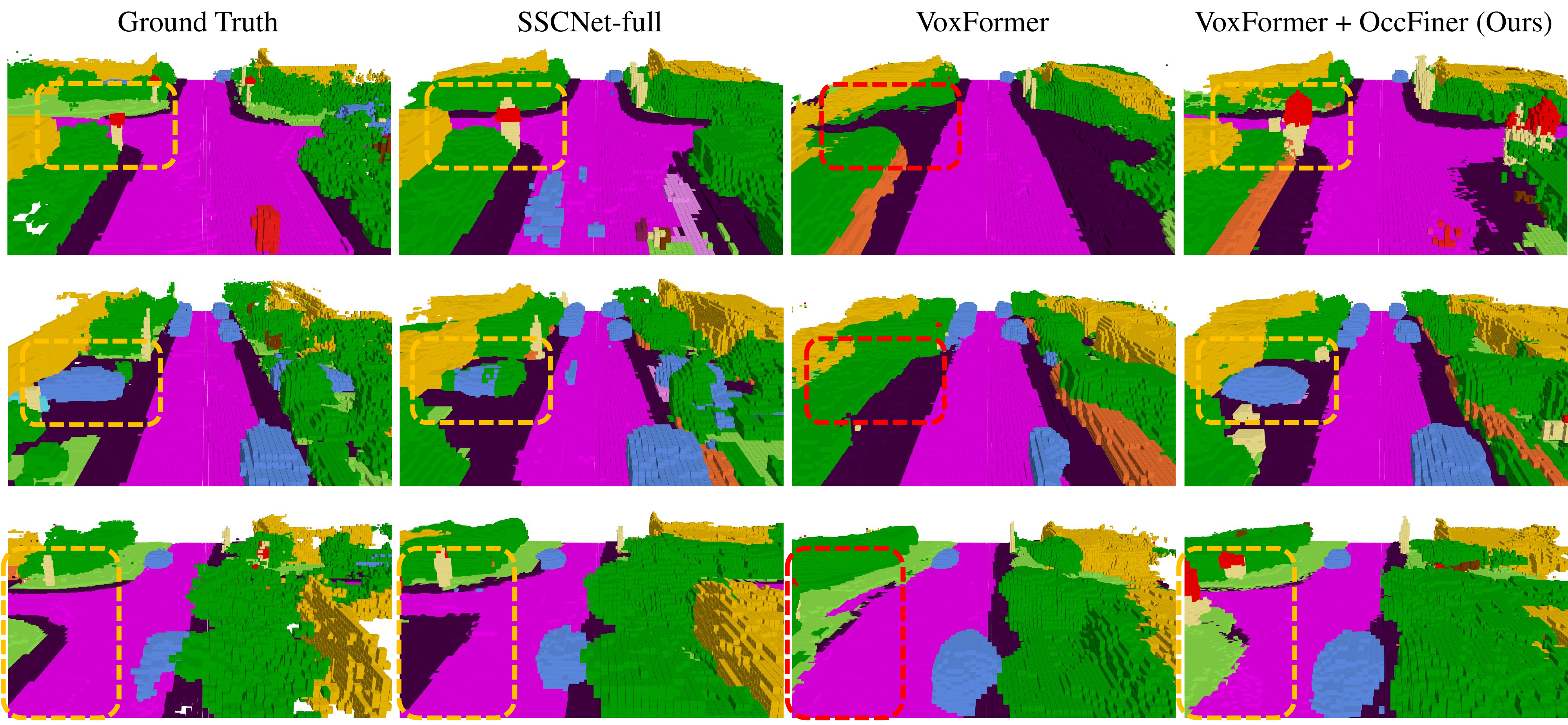}
    \vskip -2ex
    \caption{\textbf{Qualitative comparisons.} Our offboard solution effectively fixes critical errors in onboard component~\cite{li2023voxformer}, such as large areas of \emph{missing road and vehicles}. Moreover, OccFiner elevates the performance of pure visual solutions beyond the classic \emph{LiDAR-based} onboard method SSCNet-full~\cite{song2017semantic}.
    } 
    \label{fig_compare}
    \vskip -1ex
\end{figure*}

\subsection{Datasets and Implementation Details}
\label{exp:datasets_implementation}

\noindent \textbf{SemanticKITTI and SSCBench-KITTI360:}
The SemanticKITTI dataset~\cite{behley2019semantickitti} is a pivotal SSC benchmark featuring $22$ outdoor driving scenarios, segmented into train, validation, and test sets with a $10/1/11$ split. The focus is on a specific volume around the vehicle, mapped in $256 \times 256 \times 32$ voxel grids, where each voxel measures $0.2m^3$. SSCBench-KITTI360~\cite{li2023sscbench}, derived from KITTI-360~\cite{liao2022kitti}, spans $73.7km$ with extensive image and laser scan data, providing diverse geographical coverage and extended sequences. This dataset, sampled to reduce redundancy, offers around $13k$ frames, serving as a vital resource for SSC research. 
Unless otherwise specified (\eg, distance-based breakdown in Tab.~\ref{tab:comp2lidar}), all reported IoU and mIoU metrics are calculated over the standard full evaluation volume of $256 \times 256 \times 32$ voxels, corresponding to a spatial range of $51.2m \times 51.2m \times 6.4m$ centered around the ego-vehicle.

\noindent \textbf{Training and Optimization:}
Models are trained for $10$ epochs on the SemanticKITTI and $15$ on SSCBench-KITTI360, using the Adam optimizer~\cite{kingma2014adam} with a learning rate of $0.001$. The rate decay is $0.98^{epoch}$. Training is performed on $8$ RTX 3090 GPUs and includes x-y flipping augmentation in 3D volume space.

\noindent \textbf{Details for Multi-to-Multi Local Propagation Network:}
In our implementation, the Bird's-Eye View (BEV) feature and positional encoders are structured as an embedding \texttt{Linear} layer with \texttt{LayerNorm} and four convolutional blocks.
Each block uses \texttt{MaxPool} for downsampling by a factor of $2$, includes two BEV $3 \times 3$ 2D convolutions with \texttt{ReLU} activation, and outputs features with a channel of $80$. For the DualFlow4D block's BEV branch, we set the bev patch size at $W_{h}=7$, $W_{w}=7$, and $T=6$. The local and distant reference time windows are $T_{l}=4$ and $T_{r}=2$, respectively, with a reference frame sampling temporal radius of $n=10$ and a hidden dimension of $c_{e}=256$. The pillar branch of the DualFlow4D block is set at $W_{z}=32$ and $T=6$. Across all experiments, we stack $N=2$ DualFlow4D Transformer blocks to achieve the desired functionality.

\noindent \textbf{Details for Region-centric Global Propagation:}
For global propagation in each SSC frame, we use a temporal radius of $n = 25$. Camera voting weights are set at $w_{high}=1$, $w_{med}=0.1$, and $w_{low}=0.01$, with a bounding box of $25.6m \times 25.6m \times 6.4m$. For LiDAR, the voting weights range from $w_{max}=10$ to $w_{min}=0.1$. We employ vectorization to expedite the global propagation and, due to memory constraints, use single-precision (\texttt{float}) calculations for semantic voxel registration and 8-bit quantization (\texttt{uint8}) in generating city-level global street view maps.
In addition, when determining the semantic category of each voxel of the city-level SSC map, we use a three-dimensional sliding window strategy to implement \texttt{argmax} for different map chunks and finally fuse them into a global three-dimensional semantic map to further reduce memory consumption.

\input{Tabels/semantic-kitti-test}

\input{Tabels/kitti360-test}

\subsection{Comparison with State-of-the-Art Methods}
\label{exp:comparison}

As our work presents the \emph{first} offboard SSC generation, there are no existing competitors to parallel OccFiner's performance. Additionally, our OccFiner can utilize any off-the-shelf onboard model as its intermediate component. 

\noindent \textbf{Quantitatively Comparisons:} As shown in Tab.~\ref{table:kitti_val_perf}, we use the camera-based SSC algorithm~\cite{cao2022monoscene,huang2023tpvformer,li2023voxformer} as the onboard component. On the SemanticKITTI validation set~\cite{behley2019semantickitti}, OccFiner demonstrates substantial improvements in mIoU accuracy for various algorithms like MonoScene~\cite{cao2022monoscene}, TPVFormer~\cite{huang2023tpvformer}, and VoxFormer~\cite{li2023voxformer}. The mIoU accuracy is relatively improved by $21.39 \%$, $17.08 \%$, and $35.71 \%$, respectively. Notably, OccFiner (Vox) achieves the new state-of-the-art camera SSC accuracy of $18.09 \%$ mIoU. To ensure a meaningful and fair comparison, we also establish an offboard baseline algorithm, ``Average Fusion''. It performs region-centric aggregation without learning refinement or considering sensor measurement biases. Despite this, 
compared to VoxFormer + Average Fusion, 
our OccFiner (Vox) can still improve SSC quality by a large margin of over $24.59 \%$ in mIoU under the long-range setting.

\input{Tabels/auto-label}
\input{Tabels/semantickitti-global-val}
\vskip -1ex

Our exploration extends to validating OccFiner's versatility across different modalities in offboard SSC generation. As shown in Tab.~\ref{tab:comp2lidar}, for LiDAR-based SSC models like SSCNet~\cite{song2017semantic}, SSCNet-full~\cite{song2017semantic}, LMSCNet~\cite{roldao2020lmscnet}, and SCPNet~\cite{xia2023scpnet}, OccFiner consistently attains relative mIoU improvements of $36.64 \%$, $20.92 \%$, $15.65 \%$, and $14.83 \%$, respectively. This demonstrates its modality-agnostic capability. 
We observe that the enhancements offered by OccFiner are consistent across varying distances. 
This improvement is not only confined to long distances and low-accuracy scenarios but also extends to challenging high-accuracy short-range as well, indicating a robust and versatile 
perception
of the framework.
Remarkably, the accuracy of OccFiner~(VoxFormer) surpasses several widely recognized onboard LiDAR-based SSC models, including SSCNet~\cite{song2017semantic}, SSCNet-full~\cite{song2017semantic}, and LMSCNet~\cite{roldao2020lmscnet} (comparing Tab.~\ref{table:kitti_val_perf} results with Tab.~\ref{tab:comp2lidar}).
This marks a milestone where a refined camera-based occupancy system outperforms these established LiDAR-based approaches in overall mIoU.
Compared with the onboard LiDAR-based LMSCNet, the camera-based OccFiner~(VoxFormer) achieves a relative mIoU improvement of $5.24 \%$. This proves that OccFiner unleashes the potential of the camera-based SSC applications. 
This breakthrough in the camera-based versus LiDAR-based SSC model signifies a crucial advancement for autonomous driving, notably in the realms of pure visual SSC data closure and auto-labeling.

As shown in Tab.~\ref{table:kitti_test_perf}, in the results submitted for the hidden test set of SemanticKITTI~\cite{behley2019semantickitti}, OccFiner showcases remarkable performance. The OccFiner (Vox) and OccFiner~(SCPNet) establish new state-of-the-art accuracy for camera-based and LiDAR-based SSC respectively. They deliver significant relative mIoU improvements compared to the best-published results of $26.99 \%$ and $9.50 \%$, respectively. The OccFiner (Vox) consistently surpasses the SSCNet-full~\cite{song2017semantic}. This further demonstrates the efficacy and advancement of the OccFiner framework.

The effectiveness of OccFiner is also assessed on the SSCBench-KITTI360 dataset~\cite{li2023sscbench}. As shown in Tab.~\ref{tab:overall}, OccFiner~(Mono) demonstrates notable performance, achieving a relative improvement in mIoU of $7.96 \%$ compared to MonoScene~\cite{cao2022monoscene}. 
This underscores OccFiner's capability to enhance SSC accuracy across diverse datasets.

\noindent \textbf{Qualitative Comparisons:} As shown in Fig.~\ref{fig_compare}, our offboard solution, OccFiner, effectively corrects major onboard errors, notably in resolving extensive missing road and vehicle voxels. Furthermore, OccFiner enhances the effectiveness of purely visual approaches, surpassing the traditional \emph{LiDAR-based} onboard method SSCNet-full~\cite{song2017semantic} in performance, indicating its potential to advance the field of autonomous navigation with pure visual solution.

\noindent \textbf{Latency:} The online component VoxFormer~\cite{li2023voxformer} exhibits a latency of $1.52$ s/frame on a single NVIDIA RTX 3090 with single-precision computation. In contrast, the multi-to-multi local propagation network of OccFiner incurs a computational delay of $0.027$ s/frame on the same device. Furthermore, the region-centric global propagation on an Intel i9-13980HX CPU has a computational delay of 5.96 s/frame.
The latency of OccFiner's Stage 1 network primarily depends on its own architecture and the input voxel grid size. The Stage 2 latency depends on implementation, and the number of frames being aggregated (temporal window size, $n$). These timings for the VoxFormer baseline illustrate the typical computational overhead introduced by OccFiner's distinct offboard stages: a minimal increase from Stage 1 suited for near-real-time refinement if needed, and a larger latency for the comprehensive offboard aggregation in Stage 2.

\noindent 
\textbf{Analysis of Per-Class Performance:} As shown in Tab.~\ref{table:kitti_val_perf}, OccFiner~(Vox) improves overall IoU/mIoU but shows slight declines for small objects (motorcycle, poles) versus VoxFormer~\cite{li2023voxformer}. This stems from offboard multi-view temporal aggregation and temporal smoothing: although highly effective for enhancing the consistency and completeness of larger structures by leveraging consensus across frames, this aggregation may occasionally average out or fail to reinforce the sparse and potentially inconsistent predictions associated with small, thin, or infrequently observed objects, especially if the onboard model's predictions for these classes are already weak. The trade-off between scene coherence and rare-class fidelity warrants future study.

\subsection{Auto-labeling Experiments}
\label{exp:auto-label}

\begin{figure}
    \centering
    \includegraphics[width=0.5\textwidth]{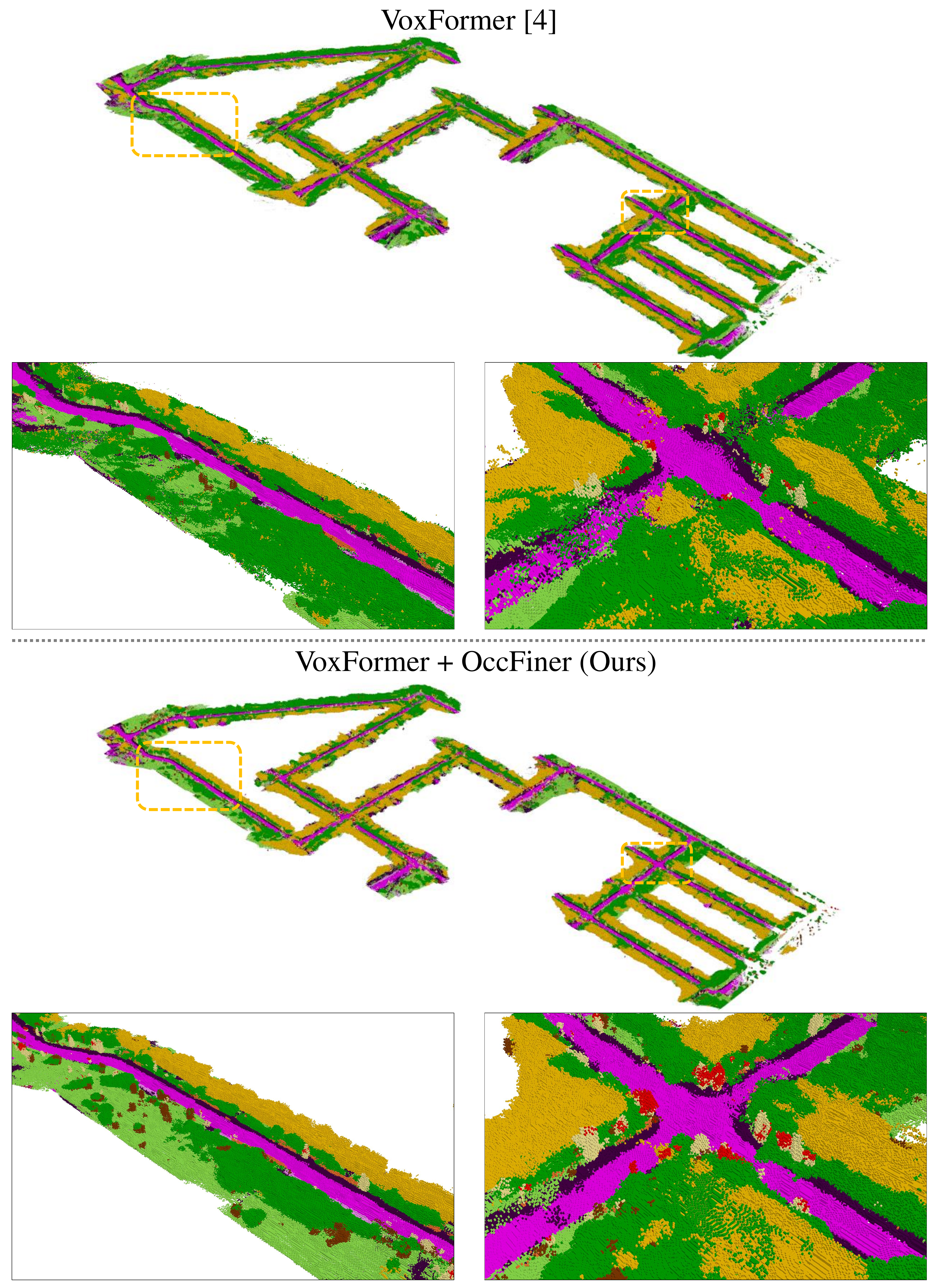}
    \caption{\textbf{City-level semantic scene completion.} The city-scale scene is expanded from the original size of $256 \times 256 \times 32$ to $2205 \times 4296 \times 261$. 
    } 
    \label{fig:shot_test}
\end{figure}

\begin{figure}
    \centering
    \includegraphics[width=0.5\textwidth]{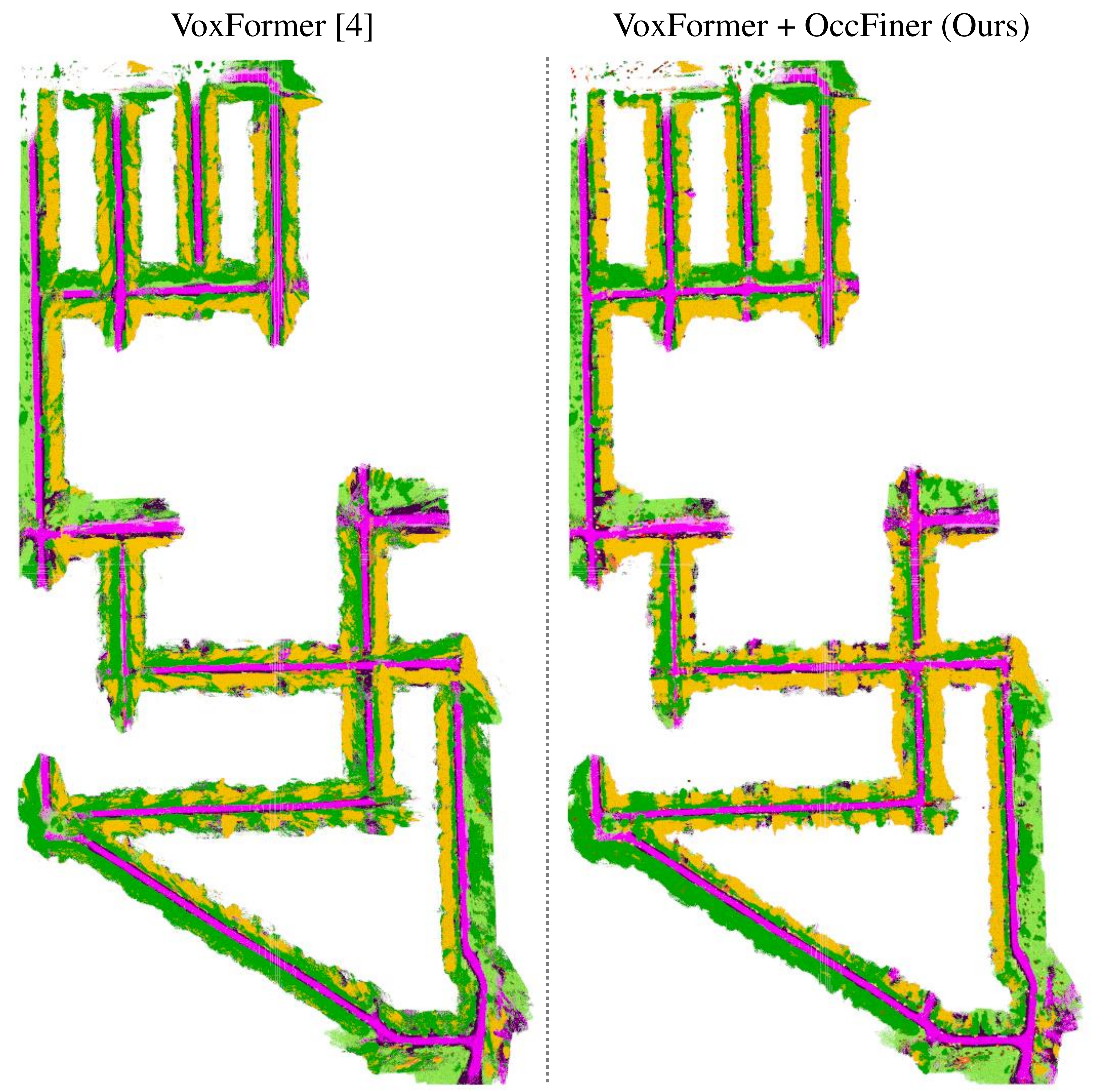}
    \caption{\textbf{City-level semantic scene completion in BEV.} Only static classes are considered. Our proposed OccFiner preserves more high-quality details and reduces camera frustum artifacts.
    } 
    \label{fig:shot_test_bev}
\end{figure}

The aforementioned results highlight OccFiner's ability to generate superior Semantic Scene Completion (SSC) labels, indicating its promise as an offboard solution for automatic SSC labeling and data closure, particularly for augmenting manual labeling in novel scenarios using pure vision-based solution. To evaluate OccFiner's label quality further, we conduct auto-labeling experiments on SemanticKITTI~\cite{behley2019semantickitti}, detailed in Table~\ref{tb:auto-label}, which simulate a real-world data loop. Utilizing manually annotated labels as a base, we enrich new scenes from the test sequences with pseudo-labels created by the pure-vision VoxFormer + OccFiner. Comparing training results of an onboard visual SSC model~\cite{zhang2023occformer} \emph{from scratch} under different labeling conditions, we find that directly using VoxFormer for `Visual Auto-Labeling' cannot achieve data closure, and the IoU and mIoU of the onboard model dropped by $0.27\%$ and $0.43\%$, respectively. Meanwhile, `LiDAR Auto-Labeling' based on SSCNet~\cite{song2017semantic} can improve the accuracy of the visual onboard model by $0.46\%$. Keeping hyperparameters and model structure consistent, we find that OccFiner's `Visual Auto-Labeling' outperforms `Human Labeling' alone. Consequently, the onboard model's IoU on the validation set improved by $0.75$\%, with a corresponding large increase of $0.73$\% in mIoU. When compared under varied labeling conditions, OccFiner's visual auto-labeling significantly improves onboard model performance, demonstrating a notable advantage over both pure human labeling and even LiDAR-based auto-labeling. This underscores our method's ability to effectively close data loops in SSC applications using high-quality, visually-derived labels.

\input{Tabels/ablations}

\subsection{City-level SSC Map Comparisons}
\label{sec:city-compare}
Previous works usually focus on local 3D semantic occupancy, \ie, onboard perception centered on a single vehicle. However, for multi-vehicle collaborative sensing, task scheduling, and traffic control, establishing a city-level SSC map is of great significance. Therefore, we propose to fuse local SSC predictions using relative poses to further explore the effectiveness of OccFiner in establishing visually derived city-level SSC results. In the global map, we only consider static rigid targets of the scene, since dynamic targets are not among the usual properties of maps. 

\noindent \textbf{Quantitative Comparisons:} As shown in Table~\ref{table:kitti_global}, we benchmark the mainstream LiDAR- and camera-based SSC methods, and explore OccFiner + VoxFormer to establish an offboard city-level SSC Map. Experimental results show that the accuracy of the LiDAR-based SSC method is consistently ahead of the camera-based method in city-level SSC maps. It is worth noting that OccFiner further improves the accuracy of both LiDAR- and camera-based city-level SSC maps. Compared with the onboard component~\cite{li2023voxformer}, the geometric IoU of OccFiner~(Vox) increased by $0.93$\%, and the semantic mIoU improved by $1.51$\%. It even exceeds the LiDAR-based SSCNet-full~\cite{song2017semantic} by $0.76$\% in geometric accuracy and surpasses the LiDAR-based SSCNet~\cite{song2017semantic} by $0.83$\% in semantic accuracy. This once again proves that the proposed OccFiner framework can effectively unleash the potential of purely visual SSC perception solutions.

\noindent \textbf{Qualitative Comparisons:} We visualize the validation sequence of SemanticKITTI~\cite{behley2019semantickitti} to evaluate the quality of city-level Semantic Scene Completion (SSC) maps. The spatial resolution of the original single-frame local occupancy is $256 \times 256 \times 32$. After splicing and voting using relative pose registration, the spatial resolution of the global occupancy map is $2205 \times 4296 \times 261$, covering a spatial range of $441 \times 859.2 \times 52.2~m^3$. As illustrated in Fig.~\ref{fig:shot_test}, directly using the outputs of the visual SSC algorithm (VoxFormer~\cite{li2023voxformer}) for global registration, despite their large onboard errors, leads to artifacts on key road surfaces. The implementation of OccFiner significantly reduces these errors, thereby enhancing the quality of the city-level SSC maps. Additionally, Fig.~\ref{fig:shot_test_bev} presents the global map visualized from a Bird's-Eye View (BEV) perspective. The VoxFormer~\cite{li2023voxformer} results exhibit notable radial artifacts, attributed to the limited viewing angle of the camera's view frustum. In contrast, OccFiner's region-centric global propagation successfully compensates for the camera's measurement limitations, effectively eliminating these radial artifacts.

\subsection{Ablation Studies}
\label{exp:ablations}

The efficacy of each module in the offboard OccFiner framework is quantified through ablation studies detailed in Tab.~\ref{table:Ablations}. 
Using VoxFormer~\cite{li2023voxformer} as the onboard component,
we evaluate mIoU accuracy across different distances. 

\noindent \textbf{Multi-to-Multi Local Propagation Network:} 
1) {DualFlow4D Transformer}. We validate the critical role of the 4D Transformer for local SSC feature propagation, evidenced by a $0.56 \%$ drop in mIoU upon its removal. 
2) {Temporal Embedding}, however, did not positively impact accuracy. 
3) {Spatial Embedding}. The result highlights the importance of spatial embedding and the introduction of a long-distance reference frame in the multi-to-multi paradigm, improving accuracy by $0.12 \%$. 
4) {Reference Frame}. OccFiner introduces a long-distance reference frame in addition to the local clip, which further improves the accuracy by $0.22 \%$. It proves the positive effect of long-term geometric and semantic cues on error compensation. 
5) {Overall}, OccFiner's first stage, while adding few model parameters ($4.51$M \textit{vs.} $551$M), effectively compensates for onboard model errors and propagates high-confidence features, boosting mIoU from $13.33 \%$ to $14.52 \%$, and uniformly improves mIoU across various distances. This is critical for the next global propagation stage.
In addition, the local propagation network can also be directly used for LiDAR-based SSC prediction. We add the SSC accuracy of the base model under various time window settings. As shown in Tab.~\ref{tb:base-model}, the accuracy consistently improves with an increase in the time window, while keeping the model structure unchanged. 
This further underscores the effectiveness and rational design of the base model structure.

\input{Tabels/base_model}

\noindent \textbf{Region-centric Global Propagation:} 
1) {Average Fusion}. Average fusion applied as a baseline to the onboard model yielded only a modest mIoU improvement at `far', indicating limitations at near and middle distances with $0.95 \%$ and $0.22 \%$ mIoU drop, respectively. 
2) {Learning Refine}. Removing the local propagation network resulted in a significant drop in mIoU by $2.04 \%$, highlighting the critical role of first-stage error compensation in enhancing global propagation accuracy across various distances. 
3) {Sensor Bias}. the frustum weighting and distance-aware weighting adjustments are shown to be vital, with their removal leading to noticeable accuracy drops. This underscores the importance of considering the camera field of view and depth uncertainties in long-distance measurements. 
4) {Temporal Input}. The optimal number of frames for global propagation is identified around $(-25, 25)$, with more frames offering no additional benefit. 5) 
{Overall}, learning-free global propagation significantly improves the accuracy of long-distance prediction. Compared with `OccFiner Stage 1 + Vanilla Registration', OccFiner's region-centric global propagation 
further improving the accuracy across different distances by $0.83 \%$, $1.10 \%$, and $1.98 \%$.

The proposed Occfiner, as the first \textit{offboard} occupancy approach, integrating both local and global propagation techniques, not only demonstrates significant improvements in predictive accuracy but also establishes a robust framework for achieving data loop-closure in intelligent transportation systems. By addressing onboard model deficiencies and enhancing error compensation through offboard processing, our methodology lays a solid foundation for future advancements in self-evolving 3D scene understanding technology, promoting safer and more reliable navigation.

%% file: Tabels/semantic-kitti-val-lidar-short-R1.tex
\begin{table*}[t]\centering
\scriptsize
\caption{\textbf{Quantitative comparison} against the state-of-the-art \textbf{LiDAR-based} SSC methods on the SemanticKITTI validation set. $\dagger$ results from the official code release. The `Relative Improvement' column shows the relative percentage improvement in the overall $51.2m \times 51.2m \times 6.4m$ mIoU achieved by OccFiner compared to its corresponding onboard baseline model listed directly above it.}
\renewcommand\tabcolsep{8pt}    
\renewcommand\arraystretch{1.5}
\begin{tabular}{l|c|ccc|ccc|c}\toprule
\multirow{2}{*}{\textbf{Methods}}  & \multirow{2}{*}{\textbf{Modality} } &\multicolumn{3}{c|}{\textbf{IoU (\%)}} &\multicolumn{3}{c|}{\textbf{mIoU (\%)}} & {\textbf{Relative} } \\
  &  &\textbf{12.8m} &\textbf{25.6m} & \textbf{51.2m} & \textbf{12.8m} & \textbf{25.6m} & \textbf{51.2m} & \textbf{Improvement} \\\midrule

\textbf{VoxFormer}~\cite{li2023voxformer}& Camera & 65.38 & 57.70 & 44.16 & 22.13 & 18.58 & 13.33 & - \\
\textbf{OccFiner (VoxFormer)
}& Camera & \textbf{65.42} & \textbf{58.50} & \textbf{47.86} & \textbf{23.43} & \textbf{21.29} & \textbf{18.09} & \textcolor{cyan}{$\uparrow$24.59\%} \\    %

\midrule

\textbf{SSCNet}~\cite{song2017semantic}& LiDAR & 50.85 &  52.54 & 44.90 & 17.59 & 17.39 & 14.11 & - \\
\textbf{OccFiner (SSCNet)}& LiDAR & \textbf{69.64} & \textbf{67.65} & \textbf{55.92} & \textbf{23.90} & \textbf{23.26} & \textbf{19.28} & \textcolor{cyan}{$\uparrow$36.64\%} \\

\midrule

\textbf{SSCNet-full}~\cite{song2017semantic}& LiDAR & {64.37} &  {61.02} & {50.22} &20.02 &  {19.68} &  {16.35} & - \\
\textbf{OccFiner (SSCNet-full)}& LiDAR & \textbf{65.96} &  \textbf{64.55} & \textbf{54.45} & \textbf{23.41} & \textbf{23.31} & \textbf{19.77} & \textcolor{cyan}{$\uparrow$20.92\%} \\

\midrule

\textbf{LMSCNet}~\cite{roldao2020lmscnet}& LiDAR & {74.88} &  {69.45} &  {55.22} &  {22.37} &  {21.50} &  {17.19} & - \\
\textbf{OccFiner (LMSC)}& LiDAR & \textbf{75.18} & \textbf{69.72} & \textbf{57.25} & \textbf{24.01} & \textbf{23.29} & \textbf{19.88} & \textcolor{cyan}{$\uparrow$15.65\%} \\

\midrule

\textbf{SCPNet$\dagger$}~\cite{xia2023scpnet}& LiDAR & 73.88 & 64.09 & 49.06 & 48.62 & 44.62 & 35.06 & - \\
\textbf{OccFiner (SCPNet)}& LiDAR & \textbf{76.44} & \textbf{67.33} & \textbf{57.19} & \textbf{48.70} & \textbf{45.90} & \textbf{40.26} & \textcolor{cyan}{$\uparrow$14.83\%} \\

\bottomrule
\end{tabular}
\label{tab:comp2lidar}
\end{table*}

%% file: Tabels/semantic-kitti-test.tex
\begin{table*}[!t]
    \scriptsize
    \setlength{\tabcolsep}{0.0035\linewidth}    
    \caption{\textbf{Semantic scene completion results on the SemanticKITTI hidden test set}~\cite{behley2019semantickitti}{.} * represents these methods are adapted for the RGB inputs, which are implemented and reported in MonoScene~\cite{cao2022monoscene}. $\dagger$ results from the official code release.}
    \resizebox{1.0\textwidth}{!}
{
    \newcommand{\classfreq}[1]{{~\tiny(\semkitfreq{#1}\%)}}  %
    \centering
    \begin{tabular}{l|c|c c c c c c c c c c c c c c c c c c c|c c}
            \toprule
    & & \multicolumn{20}{c}{} \\
    Method & Modality
    & \rotatebox{90}{\textcolor{car}{$\blacksquare$} car\classfreq{car}}
    & \rotatebox{90}{\textcolor{bicycle}{$\blacksquare$} bicycle\classfreq{bicycle}}
    & \rotatebox{90}{\textcolor{motorcycle}{$\blacksquare$} motorcycle\classfreq{motorcycle}}
    & \rotatebox{90}{\textcolor{truck}{$\blacksquare$} truck\classfreq{truck}}
    & \rotatebox{90}{\textcolor{other-vehicle}{$\blacksquare$} other-vehicle\classfreq{othervehicle}}
    & \rotatebox{90}{\textcolor{person}{$\blacksquare$} person\classfreq{person}}
    & \rotatebox{90}{\textcolor{bicyclist}{$\blacksquare$} bicyclist\classfreq{bicyclist}}
    & \rotatebox{90}{\textcolor{motorcyclist}{$\blacksquare$} motorcyclist\classfreq{motorcyclist}}
    & \rotatebox{90}{\textcolor{road}{$\blacksquare$} road\classfreq{road}}
    & \rotatebox{90}{\textcolor{parking}{$\blacksquare$} parking\classfreq{parking}}
    & \rotatebox{90}{\textcolor{sidewalk}{$\blacksquare$} sidewalk\classfreq{sidewalk}}
    & \rotatebox{90}{\textcolor{other-ground}{$\blacksquare$} other-ground\classfreq{otherground}}
    & \rotatebox{90}{\textcolor{building}{$\blacksquare$} building\classfreq{building}}
    & \rotatebox{90}{\textcolor{fence}{$\blacksquare$} fence\classfreq{fence}}
    & \rotatebox{90}{\textcolor{vegetation}{$\blacksquare$} vegetation\classfreq{vegetation}}
    & \rotatebox{90}{\textcolor{trunk}{$\blacksquare$} trunk\classfreq{trunk}}
    & \rotatebox{90}{\textcolor{terrain}{$\blacksquare$} terrain\classfreq{terrain}}
    & \rotatebox{90}{\textcolor{pole}{$\blacksquare$} pole\classfreq{pole}}
    & \rotatebox{90}{\textcolor{traffic-sign}{$\blacksquare$} traffic-sign\classfreq{trafficsign}}
    & IoU & mIoU\\
    \midrule
        LMSCNet*~\cite{roldao2020lmscnet} & Camera & 14.30 & 0.00 & 0.00 & 0.30 & 0.00 & 0.00 & 0.00 & 0.00 & 46.70 & 13.50 & 19.50 & 3.10 & 10.30 & 5.40 & 10.80 & 0.00 & 10.40 & 0.00 & 0.00 & 31.38 & 7.07 \\ 
        
        3DSketch*~\cite{chen20203d} & Camera & 17.10 & 0.00 & 0.00 & 0.00 & 0.00 & 0.00 & 0.00 & 0.00 & 37.70 & 0.00 & 19.80 & 0.00 & 12.10 & 3.40 & 12.10 & 0.00 & 16.10 & 0.00 & 0.00 & 26.85 & 6.23 \\

        AICNet*~\cite{li2020anisotropic} & Camera & 15.30 & 0.00 & 0.00 & 0.70 & 0.00 & 0.00 & 0.00 & 0.00 & 39.30 & 19.80 & 18.30 & 1.60 & 9.60 & 5.00 & 9.60 & 1.90 & 13.50 & 0.10 & 0.00 & 23.93 & 7.09 \\

        JS3C-Net*~\cite{yan2021sparse_sweep} & Camera & 20.10 & 0.00 & 0.00 & 0.80 & 4.10 & 0.00 & 0.20 & 0.20 & 47.30 & 19.90 & 21.70 & 2.80 & 12.70 & 8.70 & 14.20 & 3.10 & 12.40 & 1.90 & 0.30 & 34.00 & 8.97 \\

        OccFormer~\cite{zhang2023occformer} & Camera & {21.60} & {1.50} & \textbf{1.70} & 1.20 & 3.20 & {2.20} & 1.10 & 0.20 & {55.90} & {31.50} & {30.30} & {6.50} & {15.70} & {11.90} & {16.80} & {3.90} & {21.30} & {3.80} & {3.70} & {34.53} & {12.32} \\
        
        MonoScene~\cite{cao2022monoscene} & Camera & 18.80 & 0.50 & 0.70 & 3.30 & \textbf{4.40} & 1.00 & 1.40 & \textbf{0.40} & 54.70 & 24.80 & 27.10 & 5.70 & 14.40 & 11.10 & 14.90 & 2.40 & 19.50 & 3.30 & 2.10 & 34.16 & 11.08 \\

        TPVFormer~\cite{huang2023tpvformer} & Camera & 19.20 & 1.00 & 0.50 & {3.70} & 2.30 & 1.10 & \textbf{2.40} & 0.30 & 55.10 & 27.40 & 27.20 & {6.50} & 14.80 & 11.00 & 13.90 & 2.60 & 20.40 & 2.90 & 1.50 & 34.25 & 11.26\\

        VoxFormer~\cite{li2023voxformer} & Camera & 21.80 & 0.90 & 1.20 & 5.60 & 2.80 & 1.00 & 1.30 & 0.00 & 54.10 & 24.70 & 27.80 & 9.20 & 23.40 & 14.20 & 24.60 & 9.00 & 23.40 & \textbf{6.70} & \textbf{6.60} & 43.09 & 13.60  \\ %

        \midrule
        \rowcolor{gray!20}
        OccFiner (Vox) & Camera & \textbf{31.10} & \textbf{1.90} & 1.60 & \textbf{7.80} & 2.50 & \textbf{3.80} & 2.00 & 0.00 & \textbf{62.00} & \textbf{35.20} & \textbf{36.10} & \textbf{13.70} & \textbf{28.10} & \textbf{18.70} & \textbf{29.10} & \textbf{9.60} & \textbf{31.70} & 6.60 & 6.50 & \textbf{45.54} & \textbf{17.27} \\
        \rowcolor{gray!20}
        \textit{w.r.t. VoxFormer} & - & \red{+9.30} & \red{+1.00} & \red{+0.40} & \red{+2.20} & \green{-0.30} & \red{+2.80} & \red{+0.70} & - & \red{+7.90} & \red{+10.50} & \red{+8.30} & \red{+4.50} & \red{+4.70} & \red{+4.50} & \red{+4.50} & \red{+0.60} & \red{+8.30} & \green{-0.10} & \green{-0.10} & \red{+2.45} & \red{+3.67}\\

        \midrule
        \midrule
        SSCNet-Full~\cite{song2017semantic} & LiDAR & 24.30 & 0.50 & 0.80 & 1.20 & 4.30 & 0.30 & 0.30 & 0.00 & 51.20 & 27.10 & 30.80 & 6.40 & 34.50 & 19.90 & 35.30 & 18.20 & 29.00 & 13.10 & 6.70 & 50.00 & 16.10 \\ 
        ESSCNet~\cite{zhang2018efficient} & LiDAR & 26.40 & 0.30 & 5.40 & 5.00 & 9.10 & 2.90 & 2.70 & 0.10 & 43.80 & 26.90 & 28.10 & 10.30 & 29.80 & 23.30 & 35.80 & 20.10 & 28.70 & 16.40 & 16.70 & 41.80 & 17.50 \\
        LMSCNet~\cite{roldao2020lmscnet} & LiDAR & 30.90 & 0.00 & 0.00 & 1.50 & 0.80 & 0.00 & 0.00 & 0.00 & 64.80 & 29.00 & 34.70 & 4.60 & 38.10 & 21.30 & 41.30 & 19.90 & 32.10 & 15.00 & 0.80 & 56.70 & 17.60 \\ 
        SCPNet$\dagger$~\cite{xia2023scpnet} & LiDAR  & 42.30 & 33.70 & 33.60 & 12.20 & 26.00 & 18.40 & 17.00 & 1.60 & 69.50 & 51.70 & 49.50 & 29.20 & 35.40 & 41.50 & 41.80 & 38.40 & 49.70 & 37.80 & 27.10 & 55.15 & 34.54 \\ 

        \midrule
        \rowcolor{gray!20}
        OccFiner (SCPNet) & LiDAR & \textbf{49.00} & \textbf{38.30} & \textbf{40.20} & \textbf{13.30} & \textbf{28.10} & \textbf{19.60} & \textbf{20.00} & \textbf{1.80} & \textbf{74.70} & \textbf{52.10} & \textbf{54.70} & \textbf{30.90} & \textbf{40.90} & \textbf{44.70} & \textbf{49.20} & \textbf{40.80} & \textbf{54.40} & \textbf{38.40} & \textbf{27.40} & \textbf{61.68} & \textbf{37.82} \\
        \rowcolor{gray!20}
        \textit{w.r.t. SCPNet} & - & \red{+6.70} & \red{+4.60} & \red{+6.60} & \red{+1.10} & \red{+2.10} & \red{+1.20} & \red{+3.00} & \red{+0.20} & \red{+5.20} & \red{+0.40} & \red{+5.20} & \red{+1.70} & \red{+5.50} & \red{+3.20} & \red{+7.40} & \red{+2.40} & \red{+4.70} & \red{+0.60} & \red{+0.30} & \red{+6.53} & \red{+3.28} \\ 
        
        \bottomrule
    \end{tabular}
    }
    \\
    \label{table:kitti_test_perf}
\end{table*}

%% file: Tabels/kitti360-test.tex
\begin{table*}[t]\centering
\caption{\textbf{Benchmarking results on SSCBench-KITTI360}~\cite{li2023sscbench}. 
The default evaluation range is $51.2{\times}51.2{\times}6.4m^3$. 
}
\label{tab:overall}
\renewcommand\tabcolsep{1.2pt}
\tiny
\resizebox{1\linewidth}{!}{
\begin{tabular}{l|c|c|c| c c c c c c c c c c c c c c c c c c }
\toprule
\textbf{Method}
&\textbf{\rotatebox{90}{Input}}  &\textbf{\rotatebox{90}{IoU}} &\textbf{\rotatebox{90}{mIoU}}
      &\rotatebox{90}{\crule[carcolor]{0.13cm}{0.13cm} \textbf{car}}
      &\rotatebox{90}{\crule[bicyclecolor]{0.13cm}{0.13cm} \textbf{bicycle}}
      &\rotatebox{90}{\crule[motorcyclecolor]{0.13cm}{0.13cm} \textbf{motorcycle}}
      &\rotatebox{90}{\crule[truckcolor]{0.13cm}{0.13cm} \textbf{truck}}
      &\rotatebox{90}{\crule[othervehiclecolor]{0.13cm}{0.13cm} \textbf{other-veh.}}
      &\rotatebox{90}{\crule[personcolor]{0.13cm}{0.13cm} \textbf{person}}
      &\rotatebox{90}{\crule[roadcolor]{0.13cm}{0.13cm} \textbf{road}}  
      &\rotatebox{90}{\crule[parkingcolor]{0.13cm}{0.13cm} \textbf{parking}}
      &\rotatebox{90}{\crule[sidewalkcolor]{0.13cm}{0.13cm} \textbf{sidewalk}}
      &\rotatebox{90}{\crule[othergroundcolor]{0.13cm}{0.13cm} \textbf{other-grnd}}
      &\rotatebox{90}{\crule[buildingcolor]{0.13cm}{0.13cm} \textbf{building}}
      &\rotatebox{90}{\crule[fencecolor]{0.13cm}{0.13cm} \textbf{fence}}
      &\rotatebox{90}{\crule[vegetationcolor]{0.13cm}{0.13cm} \textbf{vegetation}}
      &\rotatebox{90}{\crule[terraincolor]{0.13cm}{0.13cm} \textbf{terrain}}
      &\rotatebox{90}{\crule[polecolor]{0.13cm}{0.13cm} \textbf{pole}}
      &\rotatebox{90}{\crule[trafficsigncolor]{0.13cm}{0.13cm} \textbf{traf.-sign}}
      &\rotatebox{90}{\crule[other-struct.color]{0.13cm}{0.13cm} \textbf{other-struct.}}
      &\rotatebox{90}{\crule[other-objectcolor]{0.13cm}{0.13cm} \textbf{other-object}}
     
\\\midrule

        \textcolor{gray}{LMSCNet~\cite{roldao2020lmscnet}} & \textcolor{gray}{L} & \textcolor{gray}{47.53} & \textcolor{gray}{13.65} & \textcolor{gray}{20.91} & \textcolor{gray}{0.00} & \textcolor{gray}{0.00} & \textcolor{gray}{0.26} & \textcolor{gray}{0.00} & \textcolor{gray}{0.00} & \textcolor{gray}{62.95} & \textcolor{gray}{13.51} & \textcolor{gray}{33.51} & \textcolor{gray}{0.20} & \textcolor{gray}{43.67} & \textcolor{gray}{0.33} & \textcolor{gray}{40.01} & \textcolor{gray}{26.80} & \textcolor{gray}{0.00} & \textcolor{gray}{0.00} & \textcolor{gray}{3.63} & \textcolor{gray}{0.00} 
        \\ 
        
        \textcolor{gray}{SSCNet~\cite{song2017semantic}} & \textcolor{gray}{L} & \textcolor{gray}{53.58} & \textcolor{gray}{16.95} & \textcolor{gray}{31.95} & \textcolor{gray}{0.00} & \textcolor{gray}{0.17} & \textcolor{gray}{10.29} & \textcolor{gray}{0.58} & \textcolor{gray}{0.07} & \textcolor{gray}{65.70} & \textcolor{gray}{17.33} & \textcolor{gray}{41.24} & \textcolor{gray}{3.22} & \textcolor{gray}{44.41} & \textcolor{gray}{6.77} & \textcolor{gray}{43.72} & \textcolor{gray}{28.87} & \textcolor{gray}{0.78} & \textcolor{gray}{0.75} & \textcolor{gray}{8.60} & \textcolor{gray}{0.67} 
        \\ 

        \midrule
        
        Voxformer~\cite{li2023voxformer} & C & \textbf{38.76} & {11.91} & {17.84} & \textbf{1.16} & {0.89}& {4.56} & {2.06}  & \textbf{1.63} & {47.01} & {9.67} & {27.21} & {2.89} & {31.18} & \textbf{4.97} & \textbf{28.99} & {14.69} & {6.51} & \textbf{6.92} & {3.79} & {2.43} 
        \\ 
        
        MonoScene~\cite{cao2022monoscene} & C & {37.87} & {12.31} & {19.34} & {0.43} & {0.58} & {8.02} & {2.03} & {0.86} & {48.35} & {11.38} & {28.13} & {3.22} & {32.89} & {3.53} & {26.15} & {16.75} & \textbf{6.92} & {5.67} & \textbf{4.20} & {3.09} 
        \\

        \midrule
        \rowcolor{gray!20}
        OccFiner (Mono) & C & 38.51 & \textbf{13.29} & \textbf{20.78} & 1.08 & \textbf{1.03} & \textbf{9.04} & \textbf{3.58} & 1.46 & \textbf{53.47} & \textbf{12.55} & \textbf{31.27} & \textbf{4.13} & \textbf{33.75} & 4.62 & 26.83 & \textbf{18.67} & 5.04 & 4.58 & 4.05 & \textbf{3.32}  \\
        \rowcolor{gray!20}
        \textit{w.r.t. MonoScene} & - & \red{+0.64} & \red{+0.98} & \red{+1.44} & \red{+0.65} & \red{+0.45} & \red{+1.02} & \red{+1.55} & \red{+0.60} & \red{+5.12} & \red{+1.17} & \red{+3.14} & \red{+0.91} & \red{+0.86} & \red{+1.09} & \red{+0.68} & \red{+1.92} & \green{-1.88} & \green{-1.09} & \green{-0.15} & \red{+0.23} \\

\bottomrule
\end{tabular}
}
\end{table*}

%% file: Tabels/auto-label.tex
\begin{table*}
    \centering
    \caption{\textbf{Auto-labeling comparison} between onboard models trained with either training set ground-truth labels (GT-Train) or auto-labeling mixed GT-Train + pseudo-labels (PL) generated by different sources. 
    }   
    \resizebox{0.8\linewidth}{!}{
    \begin{tabular}{ll|cc}
    \toprule
    \textbf{Label} & \textbf{Source} & \textbf{IoU} & \textbf{mIoU} \\ 
    \midrule
    
    GT-Train & Human Labeling & 36.42 & 13.17 \\  %

    GT-Train + PL-Test (SSCNet~\cite{song2017semantic}) & LiDAR Auto-Labeling  & 37.07 & 13.63 \\
    GT-Train + PL-Test (VoxFormer~\cite{li2023voxformer}) & Visual Auto-Labeling  & 36.15 & 12.74 \\
    GT-Train + PL-Test (VoxFormer~\cite{li2023voxformer} + OccFiner) & Visual Auto-Labeling  & \textbf{37.17} & \textbf{13.90} \\
    \bottomrule
    \end{tabular}
    }
    \label{tb:auto-label}

\end{table*}

%% file: Tabels/semantickitti-global-val.tex
\begin{table*}
    \scriptsize
    \setlength{\tabcolsep}{0.0035\linewidth}    
    \caption{\textbf{City-level semantic scene completion results on the SemanticKITTI validation set}~\cite{behley2019semantickitti}\textbf{.} Note we only compare static classes in the global map comparisons. $\dagger$ results from the official code release.}
    \vspace{-2mm}
    \resizebox{1.0\textwidth}{!}
{
    \newcommand{\classfreq}[1]{{~\tiny(\semkitfreq{#1}\%)}}  %
    \centering
    \begin{tabular}{l|c| c c c c c c c c c c c|c c}
            \toprule
    & & \multicolumn{12}{c}{} \\
    Method & Modality
    & \rotatebox{90}{\textcolor{road}{$\blacksquare$} road\classfreq{road}}
    & \rotatebox{90}{\textcolor{parking}{$\blacksquare$} parking\classfreq{parking}}
    & \rotatebox{90}{\textcolor{sidewalk}{$\blacksquare$} sidewalk\classfreq{sidewalk}}
    & \rotatebox{90}{\textcolor{other-ground}{$\blacksquare$} other-ground\classfreq{otherground}}
    & \rotatebox{90}{\textcolor{building}{$\blacksquare$} building\classfreq{building}}
    & \rotatebox{90}{\textcolor{fence}{$\blacksquare$} fence\classfreq{fence}}
    & \rotatebox{90}{\textcolor{vegetation}{$\blacksquare$} vegetation\classfreq{vegetation}}
    & \rotatebox{90}{\textcolor{trunk}{$\blacksquare$} trunk\classfreq{trunk}}
    & \rotatebox{90}{\textcolor{terrain}{$\blacksquare$} terrain\classfreq{terrain}}
    & \rotatebox{90}{\textcolor{pole}{$\blacksquare$} pole\classfreq{pole}}
    & \rotatebox{90}{\textcolor{traffic-sign}{$\blacksquare$} traffic-sign\classfreq{trafficsign}}
    & IoU & mIoU\\
    \midrule

        SSCNet~\cite{song2017semantic} & LiDAR & 32.78 & 11.88 & 18.00 & 0.00 & 14.58 & 9.17 & 26.78 & 13.48 & 26.56 & 19.88 & 4.70 & 30.06 & 16.16 \\ 

        SSCNet-full~\cite{song2017semantic} & LiDAR & 43.82 & 12.52 & 21.24 & 0.35 & 14.58 & 10.05 & 27.18 & 16.46 & 33.25 & 24.47 & 6.92 & 32.83 & 19.16 \\ 

        LMSCNet~\cite{roldao2020lmscnet} & LiDAR & 49.71 & 16.73 & 25.89 & 0.00 & 18.93 & 11.67 & 34.73 & 15.68 & 35.16 & 26.81 & 0.69 & 41.66 & 21.45 \\ 
        SCPNet$\dagger$~\cite{xia2023scpnet} & LiDAR  & \textbf{74.81} & \textbf{52.43} & \textbf{50.80} & 7.85 & 38.91 & 28.99 & 45.30 & 36.27 & 54.86 & 41.33 & 25.59 & 59.39 & 41.55 \\ 

        \rowcolor{gray!20}
        OccFiner (SCP) & LiDAR & 72.76 & 51.56 & 50.57 & \textbf{10.03} & \textbf{41.40} & \textbf{30.61} & \textbf{47.09} & \textbf{39.86} & \textbf{55.11} & \textbf{42.64} & \textbf{26.76} & \textbf{60.16} & \textbf{42.58} \\
        \rowcolor{gray!20}
        \textit{w.r.t. SCPNet} & - & \green{-2.05} & \green{-0.87} & \green{-0.23} & \red{+2.18} & \red{+2.49} & \red{+1.62} & \red{+1.79} & \red{+3.59} & \red{+0.25} & \red{+1.31} & \red{+1.17} & \red{+0.77} &\red{+1.03}\\

        \midrule
        MonoScene~\cite{cao2022monoscene} & Camera & 47.07 & 10.04 & 15.32 & 0.58 & 10.39 & 9.74 & 17.41 & 5.13 & 25.44 & 7.31 & 5.37 & 29.75 & 13.98  \\ 
        
        OccFormer~\cite{zhang2023occformer} & Camera & 46.70 & \textbf{17.90} & 14.77 & \textbf{0.95} & 9.17 & 7.31 & 20.40 & 5.93 & 24.64 & 6.87 & 5.77 & 27.63 & 14.58  \\ 
        
        TPVFormer~\cite{huang2023tpvformer} & Camera & 43.49 & 15.28 & 13.46 & 0.51 & 8.91 & 8.74 & 17.28 & 4.86 & 24.53 & 6.34 & 4.23 & 26.61 & 13.42  \\ 
        
        VoxFormer~\cite{li2023voxformer} & Camera & 42.22 & 12.48 & 15.16 & 0.36 & 13.04 & \textbf{10.12} & 22.21 & 10.02 & \textbf{25.63} & \textbf{12.36} & \textbf{6.71} & 32.66 & 15.48  \\ %

        \rowcolor{gray!20}
        OccFiner (Vox) & Camera & \textbf{47.44} & 16.22 & \textbf{21.04} & 0.83 & \textbf{13.29} & 9.96 & \textbf{27.17} & \textbf{10.62} & 24.75 & 10.33 & 5.19 & \textbf{33.59} & \textbf{16.99} \\
        \rowcolor{gray!20}
        \textit{w.r.t. VoxFormer} & - & \red{+5.22} & \red{+3.74} & \red{+5.88} & \red{+0.47} & \red{+0.25} & \green{-0.16} & \red{+4.96} & \red{+0.60} & \green{-0.88} & \green{-2.03} & \green{-1.52} & \red{+0.93} & \red{+1.51}\\

        \bottomrule
    \end{tabular}
    }
    \\
    \label{table:kitti_global}
\end{table*}

%% file: Tabels/ablations.tex
\setlength\tabcolsep{.8em}  
\begin{table*}[t]
\centering
\caption{\textbf{Ablation experiments.} Settings used in our final framework are underlined. See Sec.~\ref{exp:ablations} for details.}
\resizebox{0.85\textwidth}{!}{
\begin{tabular}{llcccr}
\toprule

\multirow{2}{*}{Experiment} & \multirow{2}{*}{Method} & \multicolumn{3}{c}{\underline{SemanticKITTI (val)}} & \multirow{2}{*}{\#Parameters} \\

& & Near & Middle & Far &  \\

\midrule 
\midrule
\rule{0pt}{1ex} \\

\multicolumn{6}{l}{\emph{Reference Model} (VoxFormer~\cite{li2023voxformer}), Training: 10 Epoch (SemanticKITTI~\cite{behley2019semantickitti}).} \\

\midrule

\multirow{1}{*}{\textcolor{blue}{Onboard Model} } 
& \textcolor{blue}{VoxFormer}~\cite{li2023voxformer} & \textcolor{blue}{22.13} & \textcolor{blue}{18.58} & \textcolor{blue}{13.33} & \textcolor{blue}{551M} \\ 

\midrule

\multirow{2}{*}{DualFlow4D Transformer} 
& \textit{w/o} & 21.71 & 19.23 & 13.96 & \textbf{+0.83M} \\ 
& \underline{\textit{with}} & \textbf{23.29} & \textbf{20.25} & \textbf{14.52} & +4.51M \\

\midrule

\multirow{2}{*}{Temporal Embedding} 
& Embedding & 23.24 & 20.21 & 14.50 & +4.90M \\
& \underline{Not Embedding} & \textbf{23.29} & \textbf{20.25} & \textbf{14.52} & \textbf{+4.51M} \\
                            
\midrule
                            
\multirow{2}{*}{Spatial Embedding} 
& Not Embedding & 23.14 & 19.95 & 14.40 & \textbf{+4.12M} \\
& \underline{Embedding} & \textbf{23.29} & \textbf{20.25} & \textbf{14.52} & +4.51M \\       %

\midrule
                            
\multirow{2}{*}{Reference Frame} 
& \textit{w/o} & 22.99 & 20.04 & 14.30 & \textbf{+4.51M} \\
& \underline{\textit{with}} & \textbf{23.29} & \textbf{20.25} & \textbf{14.52} & \textbf{+4.51M} \\

\midrule
\rule{0pt}{1ex} \\
\multicolumn{6}{l}{\emph{Reference Model} (VoxFormer~\cite{li2023voxformer}), Training: 10 Epoch (SemanticKITTI~\cite{behley2019semantickitti}) $\rightarrow$ Global Prop.} \\

\midrule
\multirow{2}{*}{\textcolor{blue}{Offboard Baseline (Average Fusion)} } & \textcolor{blue}{VoxFormer~\cite{li2023voxformer} + Vanilla Registration}  & \textcolor{blue}{21.18}  & \textcolor{blue}{18.36}  & \textcolor{blue}{14.52}  & \textcolor{blue}{551M}  \\   %
& \textcolor{blue}{OccFiner Stage 1 + Vanilla Registration}  & \textcolor{blue}{22.60}  & \textcolor{blue}{20.19}  & \textcolor{blue}{16.11}  & \textcolor{blue}{+4.51M}  \\

\midrule
\multirow{2}{*}{Learning Refine} 
& \textit{w/o} & 21.48 & 18.97 & 16.05 & \textit{n.a.} \\ 
& \underline{\textit{with}} & \textbf{23.43} & \textbf{21.29} & \textbf{18.09} & +4.51M \\  %

\midrule
\multirow{2}{*}{Frustum Weighting} 
& \textit{w/o} & 23.09 & 20.80 & 17.63 & +4.51M \\ 
& \underline{\textit{with}} & \textbf{23.43} & \textbf{21.29} & \textbf{18.09} & +4.51M \\  %

\midrule

\multirow{2}{*}{Distance-aware Weighting} 
& \textit{w/o} & 23.14 & 20.92 & 17.34 & +4.51M \\
& \underline{\textit{with}} & \textbf{23.43} & \textbf{21.29} & \textbf{18.09} & +4.51M \\  %

\midrule
                            
\multirow{6}{*}{Temporal Input}       
& (-5, 5) & 23.41 & 21.11 & 16.25 & +4.51M \\
& (-10, 10) & 23.45 & 21.29 & 17.65 & +4.51M \\
& (-15, 15) & 23.43 & 21.28 & 18.00 & +4.51M \\
& (-20, 20) & 23.43 & 21.28 & 18.05 & +4.51M \\  %
& \underline{(-25, 25)} & \textbf{23.43} & \textbf{21.29} & \textbf{18.09} & +4.51M \\  %
& (-30, 30) & 23.43 & 21.29 & 18.08 & +4.51M \\
                                        
\midrule

\end{tabular}
}
\label{table:Ablations}
\end{table*}

%% file: Tabels/base_model.tex
\begin{table}
    \centering
    \caption{Performance of \textbf{directly} using multi-to-multi propagation network with different time window. 
    }
    \resizebox{0.95\linewidth}{!}{
    \begin{tabular}{llcc}
\hline
\textbf{Dataset} & \textbf{Input} & \textbf{Time Window} & \textbf{mIoU} \\
\hline
\hline
\multirow{4}{*}{SemanticKITTI} & \multirow{4}{*}{LiDAR Scans} & 1 & 17.99 \\
& & 2 & 19.10 \\
& & 4 & 19.98 \\
& & 6 & \textbf{20.68} \\
\hline
\end{tabular}
    }
    \vspace{-5mm}
    \label{tb:base-model}
\end{table}

%% file: Tex_content/conclusion_R1.tex
\vskip -0.5ex

In addressing the challenges of inferior performance and data loop-closure in vision-based Semantic Scene Completion (SSC),
we introduce OccFiner, the first offboard SSC setup designed to enhance the reliability of onboard models. By removing computational constraints, OccFiner reasons with all frames together to build multi-view consistent SSC predictions. It employs a hybrid propagation strategy: compensating for onboard model errors and propagating semantic and geometric cues locally, while managing sensor bias and long-term information aggregation globally.
Our experiments validate OccFiner's versatility across various onboard models and its ability to significantly elevate SSC quality, setting new benchmarks on SemanticKITTI for both vision- and LiDAR-based algorithms. 
Integrating data into offboard occupancy systems successfully addresses model deficiencies, achieves data loop-closure, and enables system self-evolution. This approach also solves the distribution shift between training and deployment, allowing for scalable, cost-effective improvements in the accurate reconstruction of 3D scenes.
We hope that our framework will augment onboard SSC algorithms and inspire future offboard research, advancing autonomous driving technology by reducing manual annotation efforts and improving large-scale 3D scene understanding.

\noindent \textbf{Implications:} 
This study advances autonomous driving perception by demonstrating three key insights: (1) Vision-based systems with offboard processing (OccFiner) achieve LiDAR-comparable 3D scene understanding, challenging LiDAR's necessity for geometric reconstruction. (2) Camera-derived semantic scene completion enables scalable vision-centric systems through automated pseudo-labeling, reducing manual annotation dependence. (3) Offboard processing enhances scene representation completeness and consistency, critical for safety decisions and map automation.

\noindent \textbf{Limitations:} 
Our offboard framework, OccFiner, achieves notable results but requires individual retraining for each onboard model. This process is tailored to address the specific errors of each onboard model, with its effectiveness partly dependent on the scene's characteristics. Additionally, there is substantial scope for improving offboard SSC accuracy in comparison to the ground truth.
Furthermore, it must be acknowledged that accurate perception, especially for fine-grained geometry and semantics using only cameras, remains challenging at longer distances (\eg, towards the $51.2m$ evaluation boundary) due to inherent limitations in resolution and depth estimation, contributing to lower performance in far-range evaluations.

\noindent \textbf{Recommendations for Future Work:} Our framework demonstrates that offboard occupancy systems enable data loop-closure and mitigate distribution shifts, achieving cost-effective scalability in 3D reconstruction while reducing manual annotation burdens. Future work should optimize OccFiner's cross-model generalization, bridge residual accuracy gaps through advanced learning, and integrate refined outputs into decision systems. Ultimately, unified multitask offboard architectures could revolutionize perception. 

\noindent \textbf{Broader impact, Ethics:} 
While OccFiner improves 3D scene understanding, potential inaccuracies in its predictions could have serious implications, especially in critical applications like autonomous driving. It is crucial that such technologies be complemented by additional safety measures to mitigate risks associated with these inevitable errors.